%% file: main.tex
\documentclass{article}
\usepackage{iclr2027_conference,times}
\iclrfinalcopy
\input{math_commands.tex}

\renewcommand{\eqref}[1]{(\ref{#1})}
\usepackage{hyperref}
\usepackage{url}
\usepackage{xurl}
\usepackage{booktabs,amsmath,amssymb,amsthm,graphicx,xcolor,multirow}
\usepackage{array}
\usepackage{microtype}
\usepackage{float,placeins,tikz,enumitem,caption,longtable}
\usetikzlibrary{arrows.meta,positioning,calc,fit}
\hypersetup{colorlinks=true,linkcolor=blue!60!black,citecolor=blue!60!black,urlcolor=blue!60!black}
\newcommand{\dA}{d_{\mathcal{A}}}
\newcommand{\pub}[1]{\textcolor{black!55}{#1}}

\title{\centering
The Selection Rule Decides the Winner:\\A Pre-Registered Audit of \\Open-Set Graph Anomaly Detection}
\author{
Farhan Shahriyar Hossain$^{*,1}$\;
Taufikur Rahman Fuad$^{*,1}$\;
Md Abrar Jahin$^{*,2}$\;
Md Rizwan Parvez$^{3}$\\
\textsuperscript{1}Islamic University of Technology \\
\textsuperscript{2}University of Southern California \\
\textsuperscript{3}Qatar Computing Research Institute (QCRI) \\
\textsuperscript{*}Equal contribution (the first three authors contributed equally)
}

\begin{document}
\maketitle

\begin{abstract}
  Open-set graph anomaly detection trains on a few labeled anomalies from one class and must also find anomaly classes that were never labeled. Published results share three conventions: the test score is read at the best epoch on the test set, baseline numbers are copied from earlier papers, and most anomalies are minority classes relabeled as anomalous. We ask how much of the reported ranking these conventions decide. We re-run two recent methods, DEMO and NSReg, together with OUTPOST, a small first-order detector built for this study. All three use one protocol with identical seeds and splits on eight graphs (seven for the baselines, which cannot run on ogbn-mag), ten seeds each, and every run is scored under both the best-epoch rule and a deployable validation rule. Before the runs that test them, we registered 40 predictions. Three findings hold. First, the rule changes the leader: under the best-epoch rule, OUTPOST and NSReg each lead three of seven graphs, while under the validation rule, NSReg leads five. Second, the best-epoch bonus depends on how the benchmark was built: 0.045--0.080 AUC-ROC on the three small relabeled-class graphs and 0.002--0.014 on the three real fraud graphs. Third, pseudo-labeling in OUTPOST is worth 0.038--0.065 AUC-ROC on the same three graphs but gives no benefit on any real fraud graph. We also show that a 0.002 tie band for hyperparameter selection lies below the paired standard error on all six graphs tested, even at ten seeds. Twelve of our 40 predictions were falsified, and we report them. We close with a short reporting checklist.
\end{abstract}

\section{Introduction}\label{sec:intro}

\begin{figure}[t]
  \centering
  \begin{minipage}[t]{0.485\linewidth}\centering
    \includegraphics[width=\linewidth]{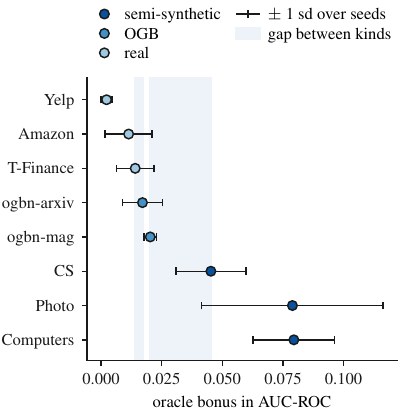}
  \end{minipage}\hfill
  \begin{minipage}[t]{0.485\linewidth}\centering
    \includegraphics[width=\linewidth]{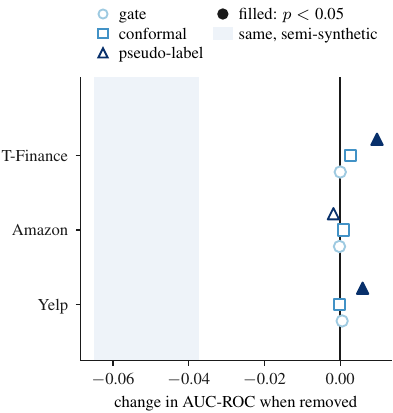}
  \end{minipage}
  \caption{Two findings of the audit, measured on OUTPOST. \textbf{Left:} the oracle bonus, the AUC-ROC gained when the reported epoch is chosen on the test set instead of on validation. Each point is one graph with a $\pm1$ standard deviation bar over ten seeds, colored by how its anomalies were produced. The bonus falls from semi-synthetic to Open Graph Benchmark (OGB) to real graphs, with no overlap among the eight means. Shaded strips mark the gaps between kinds; the gap between real and OGB graphs (0.003) lies inside the seed spread. \textbf{Right:} change in AUC-ROC when one component (atlas gate, conformal threshold, or pseudo-labeling; Section~\ref{sec:method}) is removed, on the three graphs with real fraud labels. No removal costs anything, and removing pseudo-labeling improves Yelp and T-Finance. On the semi-synthetic graphs the same removal costs 0.038 to 0.065 (shaded band).}
  \label{fig:teaser}
\end{figure}

Graph anomaly detection (GAD) identifies nodes that behave unlike the rest of a graph, such as fraudulent accounts used for money laundering \citep{ma2021survey,qiao2024survey}. An analyst can label only a few anomalies of known kinds, and kinds that appear later remain unlabeled. In open-set GAD \citep{wang2025nsreg}, a model is trained on a small set of normal nodes and anomalies from one seen class, and at test time it must rank anomalies of both seen and unseen classes above normal nodes. Recent methods report steady progress on this task \citep{wang2025nsreg,yu2026demo}.

This progress rests on three conventions that are rarely stated or tested. Reported test scores are typically read at the best epoch on the test set; we call this the \emph{oracle rule} because no deployed detector can use it. Baseline numbers are copied from earlier papers with different splits, budgets, and seed counts, which precludes paired comparison. On most benchmarks, anomalies are minority classes relabeled as anomalous; only a few benchmarks use observed fraud. Each convention is defensible alone, but together they raise a question: how much of the published ranking in open-set GAD does the evaluation protocol determine, and which conclusions survive a deployable one?

We address this question by applying one protocol to every method (Section~\ref{sec:protocol}). We re-run DEMO \citep{yu2026demo} and NSReg \citep{wang2025nsreg} from their released code alongside OUTPOST, a small first-order detector built for this study (Section~\ref{sec:method}), on eight graphs with identical splits and seeds, using ten seeds and 400 epochs for each method. Every run is scored twice: under the oracle rule, and under a \emph{validation rule} that reads the test metric at the epoch of highest validation AUC-ROC (area under the receiver operating characteristic curve), the rule a practitioner can apply. Three of the eight graphs carry observed fraud labels rather than relabeled classes. Before the runs that test them, we committed 40 predictions to version control, each with its falsifying outcome \citep{nosek2018prereg}.

OUTPOST serves two purposes. It is a competitive detector in its own right, and it is the instrument through which we exercise the protocol, since measuring a component's contribution by removing it requires a method one controls. It pairs a two-layer GraphSAGE encoder with a prototype model of normality, conformal pseudo-labeling, and similarity-ordered neighbor sampling. It requires neither a second-order gradient nor a dense propagation matrix, and carries $0.02$ to $0.52$ times the parameters of DEMO. As a result, it is the only one of the three that runs on all eight graphs: DEMO's dense personalized-PageRank matrix and NSReg's regularizer over all ordered pairs of labeled normal nodes are infeasible at the scale of ogbn-mag. OUTPOST has the highest mean AUC-ROC under both selection rules on CS and Yelp, the only graph in the study that combines observed fraud labels with a published field.

Our main contributions are as follows. (i)~We compare three open-set detectors, two of them re-run from released code, on eight graphs under one pre-registered protocol with identical splits and seeds, scoring every run under both the oracle and the validation rule, and find that the rule decides which method leads on two of seven graphs. (ii)~We show that benchmark provenance governs the size of this effect: the oracle bonus, the gain of the oracle rule over the validation rule, is $0.045$ to $0.080$ AUC-ROC on semi-synthetic graphs against $0.002$ to $0.014$ on graphs with observed fraud (Fig.~\ref{fig:teaser}, left), and on the former it exceeds 13 of the 14 paired differences between methods under the same rule. (iii)~We introduce OUTPOST, a first-order detector that scales to ogbn-mag, where neither baseline runs, and use it to show that pseudo-labeling carries the gains on semi-synthetic benchmarks and contributes nothing on any graph with real fraud (Fig.~\ref{fig:teaser}, right).

\section{Related Work}\label{sec:related}

\textbf{Graph anomaly detection.} GAD methods identify anomalous nodes through reconstruction \citep{ding2019dominant,fan2020anomalydae}, contrastive learning \citep{liu2021cola,xu2022conad}, or one-class modeling of normality \citep{wang2021ocgnn,qiao2023tam}. With limited supervision, ConsisGAD uses consistency training \citep{chen2024consisgad}, GGAD generates outliers from labeled normal nodes \citep{qiao2024ggad}, and SpaceGNN learns across multiple geometric spaces \citep{dong2025spacegnn}. These methods address label scarcity, but open-set evaluation adds a requirement they were not designed for: anomaly classes absent from the labeled training set must also be detected.

\textbf{Open-set GAD and detector design.} Open-set recognition addresses classes unseen during training \citep{scheirer2013openset,bendale2016openmax}, and open-set supervised anomaly detection extends this concern to unseen anomaly types \citep{ding2022swans}. NSReg regularizes graph representations with normal-node-oriented relations to separate normal nodes from unseen anomalies \citep{wang2025nsreg}. DEMO combines anomaly mixup, energy-gradient reweighting, and history-guided, class-adaptive pseudo-labeling \citep{yu2026demo}. OUTPOST builds on GraphSAGE \citep{hamilton2017graphsage}, conformal calibration \citep{vovk2005conformal,angelopoulos2023conformal}, and consistency-based pseudo-labeling \citep{sohn2020fixmatch,zhang2021flexmatch}; its neighbor sampling follows fraud detectors that select or reweight neighbors \citep{dou2020caregnn,liu2021pcgnn,gao2023ghrn}, with raw-feature similarity in place of a learned selector.

\textbf{GAD benchmarks and anomaly provenance.} BOND compares unsupervised detectors on synthetic and organic outliers \citep{liu2022bond}, and GADBench evaluates fully supervised and limited-label detectors on organic anomalies, with standardized tuning and model selection by validation AUC-PR (area under the precision--recall curve) \citep{tang2023gadbench}. Both show that benchmark construction and evaluation protocol shape comparisons, but neither covers the open-set setting. We contrast the construction most open-set benchmarks use, minority classes relabeled as anomalies rather than injected anomalous edges or attributes, with observed fraud labels.

\textbf{Model selection and reproducible comparison.} Re-evaluations of node and graph classification show that data splits, training procedures, and benchmark design can change conclusions about progress in graph neural networks \citep{shchur2018pitfalls,errica2020fair,platonov2023critical}. DomainBed compares model-selection strategies for generalization to unseen domains \citep{gulrajani2021lost}, and studies of benchmark variance and tuning budgets show that small reported gains require controlled comparisons \citep{bouthillier2021variance,dodge2019show}. To our knowledge, open-set GAD has had no such re-evaluation. We score each run under both the oracle rule and a validation rule that uses only seen-class anomalies, so the effect of checkpoint choice on rankings and on component effects is measured within the same runs.

\section{Problem Setting and Evaluation Protocol}\label{sec:protocol}

\textbf{Task.} Let $G=(V,E,X)$ be an attributed graph with node features $x_v\in\mathbb{R}^{d_0}$, where $d_0$ is the input feature dimension. The nodes split into normal nodes $V_n$ and anomalies $V_a$, with $|V_a|\ll|V_n|$, and $V_a$ is a union of anomaly classes. Training uses labeled normal nodes and labeled anomalies from a single \emph{seen} class. A detector outputs a score $s(v)\in[0,1]$ and should satisfy $s(v_a)>s(v_n)$ for anomalies of every class, seen or unseen. As in prior work, the setting is transductive: the whole graph is visible during training, and the unlabeled pool used for self-training is the test node set, whose labels no method reads. The splits are disjoint and the pool is shared by all three methods, so transductive access is a property of the task, not an advantage of one detector.

\textbf{Graphs.} We use eight graphs of three kinds (Table~\ref{tab:datasets}, Appendix~\ref{app:data}). \emph{Semi-synthetic}: Photo, Computers, and CS \citep{shchur2018pitfalls}, where every class holding at most 5\% of the nodes is relabeled as an anomaly class. \emph{OGB}: ogbn-arxiv and the paper graph of ogbn-mag \citep{hu2020ogb}, relabeled in the same way with class-size bands of 3--5\% and at most 0.03\%, respectively. \emph{Real}: Yelp \citep{rayana2015yelp}, Amazon \citep{mcauley2013amazon}, and T-Finance \citep{tang2022bwgnn}, whose anomaly labels are observed fraud. These graphs are binary and have no unseen class; they test the few-label regime with real anomalies, and open-set claims rest on the five multi-class graphs.

\textbf{Splits and rotations.} Following \citet{wang2025nsreg} and \citet{yu2026demo}, training uses 50 anomalies of the seen class and 5\% of the normal nodes; validation uses 30 anomalies of the same class and 1\% of the normal nodes; all other nodes, including every unseen-class anomaly, are test nodes. Each anomaly class takes one turn as the seen class (a \emph{rotation}), and a graph's score is the mean over its rotations. A seed fixes both the split and the initialization, so the spread over seeds covers both.

\textbf{Two selection rules.} Let $m_t$ be a test metric after epoch $t$ and $a_t$ the validation AUC-ROC. The \emph{oracle rule} reports $\max_t m_t$, separately for each metric; to our knowledge, the published numbers we compare with use this rule. The \emph{validation rule} reports $m_{t^\star}$ with $t^\star=\arg\max_t a_t$; it reads no test label and can be deployed. We call their difference, $\max_t m_t-m_{t^\star}$, the \emph{oracle bonus}. The validation rule alone would understate earlier work, whose numbers were not produced under it, and the oracle rule alone would overstate what these detectors deliver in use, so we record both and read the validation rule as the deployment estimate. We report AUC-ROC, the headline metric of prior work, and AUC-PR, which is more informative under strong imbalance \citep{davis2006prroc}.

\textbf{Baselines under the same protocol.} NSReg defines the protocol we inherit, and DEMO holds the best published AUC-ROC on every graph for which one exists, so we re-run both. DEMO runs in its published configuration with mixup enabled, omitting only its second-order energy term, which adds 0.001 AUC-ROC on Photo ($p=0.83$) at 6.5$\times$ the compute; NSReg runs from its released code. The code changes needed to run DEMO fix a crash, grant it more budget, or leave its metrics untouched (Appendix~\ref{app:baselines}). Both receive our graph tensors, our splits (a hash of each split is stored with each run and matched across methods), the same ten seeds, the same metrics, and both rules. All methods train for 400 epochs, the budget at which our DEMO re-run is properly trained, although this budget costs OUTPOST a tie on Photo (Appendix~\ref{app:budget}); ablations run at OUTPOST's development budget, 200 epochs on the semi-synthetic graphs. They also receive the tuning rule applied to OUTPOST; the only selection that survived ten seeds strengthens NSReg on Amazon, and we adopt it. Numbers for 16 earlier methods are transcribed from \citet{yu2026demo} and shown separately (Appendix~\ref{app:published}), since values from other splits and budgets cannot enter a paired test.

\textbf{Statistics.} Table cells give the mean and sample standard deviation over seeds. We pair comparisons between re-run methods by seed, test them with a two-sided Wilcoxon signed-rank test \citep{wilcoxon1945,demsar2006statistical}, and also report wins and losses over seeds. With $n$ pairs the smallest attainable $p$-value is $2^{-(n-1)}$, which is 0.0625 for $n=5$, so we fixed $n=10$ for every main comparison before seeing the added runs, lowering the floor to 0.002. We apply Holm's correction within ablation families \citep{holm1979} and treat any comparison with a transcribed value as weak evidence.

\textbf{Pre-registration.} We committed each of 40 predictions (P1--P40), with its falsifying outcome, before the runs that test it. The ledger records 17 confirmed, 12 falsified, 3 split, and 1 null; the other 7 are superseded, corrected, or partial (Appendix~\ref{app:prereg}). We cite predictions by number where they matter.

\section{OUTPOST: A First-Order Open-Set Detector}\label{sec:method}

\begin{figure}[tb]
  \centering
  \includegraphics[width=\linewidth]{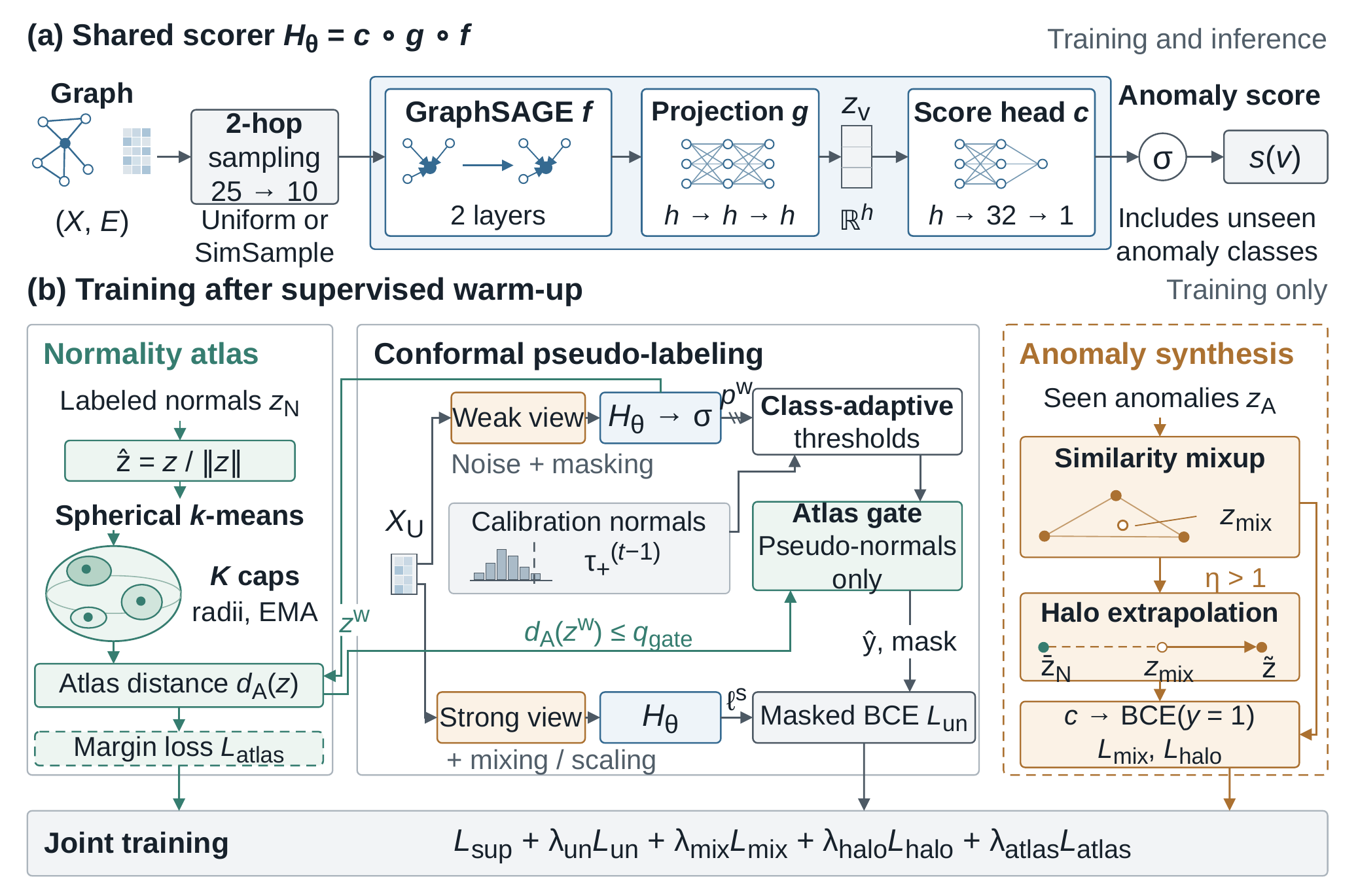}
  \caption{Overview of OUTPOST. \textbf{(a)}~The scorer $H_\theta=c\circ g\circ f$ shared by training and inference: a GraphSAGE encoder $f$, a projection $g$, and a score head $c$ over a sampled two-hop neighborhood; $s(v)=\sigma(c(z_v))$ alone ranks test nodes, including those of never-labeled anomaly classes. \textbf{(b)}~Training after the supervised warm-up. The normality atlas fits $K$ spherical caps to labeled normals and measures the distance $\dA$ to them. Conformal pseudo-labeling recalibrates the positive threshold $\tau_+$ on held-out normals every epoch and accepts a pseudo-normal only inside the atlas. Anomaly synthesis adds positives by similarity mixup and halo extrapolation. Dashed modules run only on relabeled-class graphs; on the three graphs with observed fraud, synthesis and $\mathcal{L}_{\text{atlas}}$ are off and only the atlas gate remains. $h$ is the hidden width, $\bar z_N$ the detached mean of normal embeddings, $q_{\text{gate}}$ a quantile of $\dA$ over labeled normals, and $/\!/$ a stopped gradient.}
  \label{fig:method}
\end{figure}

OUTPOST is designed around three requirements, two from the setting of Section~\ref{sec:protocol} and one from the audit. Supervision covers a single anomaly class, so the model must describe normality well enough that an unlabeled class falls outside it. With $736{,}389$ nodes in the largest graph, no component may rest on a dense propagation matrix or a second-order gradient. And each component must be removable on its own, so that its contribution can be measured. Figure~\ref{fig:method} shows the five components that follow from these requirements; Appendix~\ref{app:hparams} gives all settings.

\textbf{Encoder and score.} For a target node $v$ we sample a two-hop computation graph $\mathcal{N}(v)$ with fan-out $(25,10)$, as in GraphSAGE \citep{hamilton2017graphsage}. A two-layer GraphSAGE encoder $f$ and a two-layer projection $g$ give the embedding $z_v=g(f(x,\mathcal{N}(v)))\in\mathbb{R}^{h}$, with $h$ the hidden width, and a scoring head $c$ gives the anomaly score
\begin{equation}
  s(v)=\sigma\big(c(z_v)\big),\qquad \mathcal{L}_{\text{sup}}=\mathrm{BCE}\big(c(z_v),y_v\big)\ \text{over labeled } v, \label{eq:score}
\end{equation}
where $\sigma$ is the sigmoid, $\mathrm{BCE}$ the binary cross-entropy, and $y_v\in\{0,1\}$. Equation~\eqref{eq:score} is the only score used at test time (Fig.~\ref{fig:method}a); the atlas shapes training but never scores a node. We inherit two conventions from the baselines so that the comparison does not turn on them: the loss accumulates over all batches with one optimizer step per epoch, and neighbor sampling stays active at evaluation.

\textbf{Atlas of normality.} A detector supervised on one anomaly class can recognize an unseen class only as a departure from normality, which is why we model normality rather than the seen anomalies. Normal nodes need not occupy a single region, so we represent normality on the unit sphere as a union of spherical caps. At the end of the $W$ warm-up epochs, spherical $k$-means on the normalized embeddings $\hat z=z/\lVert z\rVert$ of labeled normal nodes gives $K$ prototypes $\mu_j$. The radius $r_j$ is the $(1-\alpha_r)$ quantile of the geodesic distance $\arccos(\hat z^{\top}\mu_j)$ over the normals assigned to $\mu_j$; prototypes and radii then follow an exponential moving average. The distance of a node to the atlas is
\begin{equation}
  \dA(z)=\min_{1\le j\le K}\big[\arccos(\hat z^{\top}\mu_j)-r_j\big]_+ , \label{eq:atlas}
\end{equation}
which is zero inside any cap. Equation~\eqref{eq:atlas} enters the model twice: through an auxiliary loss $\mathcal{L}_{\text{atlas}}=\mathbb{E}_{\text{normal}}[\dA]+\mathbb{E}_{\text{anom}}[m-\dA]_+$ with margin $m$, and through the gate described next.

\textbf{Conformal pseudo-labeling with an atlas gate.} After warm-up, each unlabeled node $v\in U$ receives a weak view $x^{w}$ (small noise and feature masking) and a strong view $x^{s}$ (plus feature mixing and scaling), as in FixMatch \citep{sohn2020fixmatch}. Let $p_v=\sigma(c(z^{w}_v))$, computed without gradient, and $\hat y_v=\mathbf{1}[p_v\ge 0.5]$. Write $\phi(\beta)=\beta/(2-\beta)$ for the class-progress map of FlexMatch \citep{zhang2021flexmatch}, where $\beta_c\in[0,1]$ is the learning progress of class $c$. We accept a pseudo-label when
\begin{equation}
  \big(\hat y_v=1\ \wedge\ p_v\ge \phi(\beta_1)\,\tau_+\big)\quad\text{or}\quad\big(\hat y_v=0\ \wedge\ p_v\le\big(2-\phi(\beta_0)\big)\tau_-\ \wedge\ \dA(z^{w}_v)\le q_{\text{gate}}\big). \label{eq:accept}
\end{equation}
DEMO fixes $\tau_+=0.95$, a threshold whose false-positive rate depends on the score distribution and therefore on the graph. We set it instead by split conformal calibration \citep{vovk2005conformal} at every epoch: with $n$ validation normals and sorted scores $p_{(1)}\le\dots\le p_{(n)}$, $\tau_+=p_{(\lceil (n+1)(1-\alpha_+)\rceil)}$, clipped to $[0.5,0.995]$. Under exchangeability a normal node then exceeds $\tau_+$ with probability at most $\alpha_+$, so the rate of false pseudo-anomalies is controlled rather than assumed. The normal-side threshold $\tau_-$ stays fixed, because the gate catches false pseudo-normals instead. The \emph{atlas gate} is the last condition of Eq.~\eqref{eq:accept}: a node may be pseudo-labeled normal only if it lies inside the atlas, with $q_{\text{gate}}$ the 0.9 quantile of $\dA$ over labeled normals in the current epoch. It guards against a scattered anomaly, whose score low-pass propagation has drawn toward its normal neighbors, being confirmed as normal and then reinforced. The loss is $\mathcal{L}_{\text{un}}=\frac{1}{|U|}\sum_{v\in U}\mathbf{1}[\text{Eq.~\eqref{eq:accept}}]\,\mathrm{BCE}(c(z^{s}_v),\hat y_v)$.

\textbf{Synthetic anomalies.} A single labeled class covers only part of the anomaly space, so on the five relabeled-class graphs we extend coverage in embedding space with two kinds of synthetic positive. \emph{Mixup} \citep{zhang2018mixup,yu2026demo} replaces each labeled anomaly by a softmax-similarity-weighted average of the other labeled anomalies. \emph{Halo} pushes that mixed embedding $z$ away from the mean normal embedding $\bar z_N$: $\tilde z=\bar z_N+\eta\,(z-\bar z_N)$ with $\eta\sim\mathcal{U}[1.2,2.0]$, covering regions beyond the seen class; with mixup off it acts on the labeled anomaly itself. Both are trained with label 1 through Eq.~\eqref{eq:score}, giving $\mathcal{L}_{\text{mix}}$ and $\mathcal{L}_{\text{halo}}$. DEMO draws its mixup weights from a dense $|V|\times|V|$ personalized-PageRank matrix, which is 115\,GB on ogbn-arxiv and 2.17\,TB on ogbn-mag (Appendix~\ref{app:baselines}); our weights are quadratic in the 50 labeled anomalies and need no such matrix.

\textbf{SimSample.} On dense fraud graphs a uniform sample of 25 neighbors is dominated by the normal neighbors of a fraud node. SimSample sorts each node's neighbor list once by the cosine similarity of raw features. At a hop with budget $b$, a node of degree $d$ then draws $k=\min(d,b)$ neighbors: the first $\lfloor\rho k\rfloor$ in similarity order and the rest uniformly. It has no parameters, uses no labels, and adds no cost at sampling time, since the ordering is built once. We set $\rho=1$ on Yelp, choose $\rho\in\{0,1\}$ on validation for Amazon and T-Finance, and set $\rho=0$ elsewhere. Section~\ref{sec:rq5} separates its effect from sampling randomness with a placebo, a dose--response curve, and an intervention on $b$.

\textbf{Full objective.} With loss weights $\lambda$ and indicator $\mathbf{1}_{t\ge W}$ for the epochs after warm-up,
\begin{equation}
  \mathcal{L}=\mathcal{L}_{\text{sup}}+\mathbf{1}_{t\ge W}\big(\lambda_{\text{un}}\mathcal{L}_{\text{un}}+\lambda_{\text{mix}}\mathcal{L}_{\text{mix}}+\lambda_{\text{halo}}\mathcal{L}_{\text{halo}}+\lambda_{\text{atlas}}\mathcal{L}_{\text{atlas}}\big). \label{eq:loss}
\end{equation}
Every term in Eq.~\eqref{eq:loss} is first order, which is what lets OUTPOST run on ogbn-mag, where neither baseline can. It has 5{,}953 to 889{,}793 parameters, 0.019--0.516$\times$ those of DEMO and 0.246--0.995$\times$ those of NSReg, and peak GPU memory 0.34--0.87$\times$ that of DEMO (Appendix~\ref{app:efficiency}). To limit the tuning decisions the protocol must account for, configurations are fixed per graph kind rather than per graph: the five relabeled-class graphs share one, and the three real graphs share another, in which the synthesis terms and $\mathcal{L}_{\text{atlas}}$ are off and only the gate remains. Amazon and T-Finance inherit the Yelp configuration untuned. Two further variants, a spectral gate and a per-node fusion with a propagation-free view, were tested and rejected (Appendix~\ref{app:ablations}).

\section{Results}\label{sec:experiments}

\begin{table}[t]
  \centering
  \caption{AUC-ROC over all test anomalies, as mean and standard deviation over ten seeds. Each run is scored at the best epoch on the test set (oracle rule) and at the epoch with the highest validation AUC-ROC (validation rule). On each graph and under each rule, the best of the three methods is bold and the runner-up underlined; the leader changes with the rule on Photo and Computers. The published column, in gray, is the best of sixteen earlier methods as reported by \citet{yu2026demo}, from other splits and budgets, and is shown for reference only. Neither baseline runs on ogbn-mag, so its row is unmarked (Appendix~\ref{app:baselines}). AUC-PR is in Table~\ref{tab:mainpr}.}
  \label{tab:main}
  \small
  \setlength{\tabcolsep}{5pt}
  \begin{tabular}{@{}lccccccc@{}}
    \toprule
    & \pub{Best} & \multicolumn{3}{c}{Best epoch on test} & \multicolumn{3}{c}{Best epoch on validation} \\
    \cmidrule(lr){3-5}\cmidrule(lr){6-8}
    Graph & \pub{published} & DEMO & NSReg & OUTPOST & DEMO & NSReg & OUTPOST \\
    \midrule
    \multicolumn{8}{@{}c}{\emph{Semi-synthetic}} \\
    Photo & \pub{0.902} & \textbf{0.888}\,{\scriptsize$\pm$.015} & \underline{0.876}\,{\scriptsize$\pm$.041} & 0.870\,{\scriptsize$\pm$.024} & 0.774\,{\scriptsize$\pm$.060} & \textbf{0.793}\,{\scriptsize$\pm$.063} & \underline{0.791}\,{\scriptsize$\pm$.048} \\
    Computers & \pub{0.844} & 0.778\,{\scriptsize$\pm$.015} & \underline{0.837}\,{\scriptsize$\pm$.023} & \textbf{0.851}\,{\scriptsize$\pm$.010} & 0.727\,{\scriptsize$\pm$.013} & \textbf{0.785}\,{\scriptsize$\pm$.022} & \underline{0.771}\,{\scriptsize$\pm$.017} \\
    CS & \pub{0.945} & 0.965\,{\scriptsize$\pm$.004} & \underline{0.970}\,{\scriptsize$\pm$.004} & \textbf{0.984}\,{\scriptsize$\pm$.002} & 0.901\,{\scriptsize$\pm$.019} & \underline{0.929}\,{\scriptsize$\pm$.017} & \textbf{0.939}\,{\scriptsize$\pm$.014} \\
    \addlinespace[2pt]
    \multicolumn{8}{@{}c}{\emph{OGB}} \\
    ogbn-arxiv & \pub{0.636} & 0.618\,{\scriptsize$\pm$.011} & \textbf{0.653}\,{\scriptsize$\pm$.010} & \underline{0.623}\,{\scriptsize$\pm$.008} & 0.604\,{\scriptsize$\pm$.007} & \textbf{0.630}\,{\scriptsize$\pm$.008} & \underline{0.606}\,{\scriptsize$\pm$.008} \\
    ogbn-mag & \pub{0.497} & -- & -- & 0.593\,{\scriptsize$\pm$.003} & -- & -- & 0.573\,{\scriptsize$\pm$.003} \\
    \addlinespace[2pt]
    \multicolumn{8}{@{}c}{\emph{Real}} \\
    Yelp & \pub{0.710} & 0.731\,{\scriptsize$\pm$.011} & \underline{0.739}\,{\scriptsize$\pm$.014} & \textbf{0.745}\,{\scriptsize$\pm$.016} & \underline{0.725}\,{\scriptsize$\pm$.013} & \underline{0.725}\,{\scriptsize$\pm$.015} & \textbf{0.743}\,{\scriptsize$\pm$.017} \\
    Amazon & \pub{--} & \underline{0.956}\,{\scriptsize$\pm$.005} & \textbf{0.959}\,{\scriptsize$\pm$.008} & 0.953\,{\scriptsize$\pm$.005} & \underline{0.951}\,{\scriptsize$\pm$.008} & \textbf{0.954}\,{\scriptsize$\pm$.008} & 0.942\,{\scriptsize$\pm$.009} \\
    T-Finance & \pub{--} & \underline{0.916}\,{\scriptsize$\pm$.008} & \textbf{0.924}\,{\scriptsize$\pm$.009} & 0.909\,{\scriptsize$\pm$.011} & \underline{0.903}\,{\scriptsize$\pm$.008} & \textbf{0.916}\,{\scriptsize$\pm$.009} & 0.895\,{\scriptsize$\pm$.013} \\
    \bottomrule
  \end{tabular}
\end{table}
\subsection{The Selection Rule Changes the Leader}\label{sec:rq1}
Table~\ref{tab:main} scores three methods on identical splits and seeds, under two rules that differ only in the epoch at which the test metric is read. Under the oracle rule, OUTPOST leads Computers, CS, and Yelp; NSReg leads Amazon, T-Finance, and ogbn-arxiv; and DEMO leads Photo. Under the validation rule, NSReg leads five graphs, OUTPOST two, and DEMO none. The leader thus changes on two graphs. On Computers, OUTPOST is ahead of NSReg by $0.0140$ under the oracle rule, on nine of ten seeds ($p=0.014$), and behind by the same margin under the validation rule. On Photo, its difference from DEMO moves from $-0.0176$ to $+0.0170$, neither significant, so the rule rather than the methods fixes the sign of that comparison. Scoring only unseen classes moves the leader on two of the four graphs where all three methods run (Tables~\ref{tab:unseen} and~\ref{tab:rankunseen}, Appendix~\ref{app:paired}).

\begin{figure}[t]
  \centering
  \begin{minipage}[c]{0.40\linewidth}\centering
    \includegraphics[width=\linewidth]{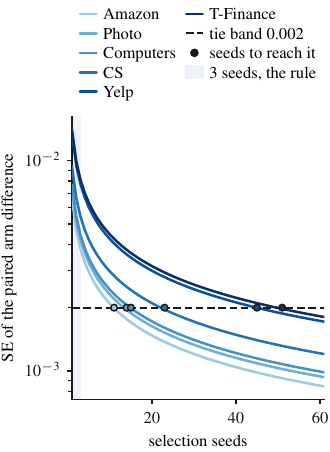}
  \end{minipage}\hfill
  \begin{minipage}[c]{0.58\linewidth}\centering
    \includegraphics[width=\linewidth]{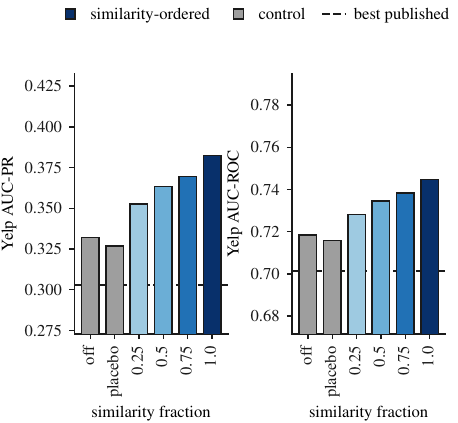}
  \end{minipage}
  \caption{\textbf{Left:} standard error of the paired validation difference between two arms against the number of seeds, one curve per graph. The dashed line is the 0.002 tie band. At the three seeds the selection rule used, every graph lies above the band; the marked point on each curve is the seed count at which it first falls inside (11 to 51). \textbf{Right:} SimSample on Yelp. Both metrics rise with the similarity fraction $\rho$. The placebo keeps the deterministic sampling but shuffles the order and lands below the arm with SimSample off, so the gain comes from which neighbors are chosen, not from reduced sampling randomness. The dashed line is the best published Yelp result.}
  \label{fig:noise}
\end{figure}

No method leads under both rules on more than three of the seven comparable graphs. The leads that survive a change of rule are OUTPOST's on CS and Yelp and NSReg's on Amazon, T-Finance, and ogbn-arxiv. In AUC-PR (Table~\ref{tab:mainpr}), OUTPOST leads Computers, CS, and Yelp under both rules. On ogbn-mag, where neither baseline can run, OUTPOST reaches $0.593$ against a best published value of $0.497$, a margin of $0.096$ on ten of ten seeds. The reference is transcribed, so this comparison is unpaired (Section~\ref{sec:protocol}); no baseline can be re-run on this graph.

Transcribed numbers and re-runs disagree in both directions, so a table mixing the two cannot be read as a ranking. Our DEMO re-run lies $0.066$ below its published AUC-ROC on Computers and $0.020$ above it on CS, and its published Yelp AUC-PR of $0.2238$ becomes $0.3504$; NSReg re-runs $0.037$ to $0.097$ above its published values on all four graphs that have one (Table~\ref{tab:repro}; Fig.~\ref{fig:yelp} shows the Yelp column). Because neither direction is systematic, a transcribed baseline is not a lower bound on what the method achieves under a common protocol.

\textbf{Hyperparameter selection.} The same problem arises in a sharper form. All three methods received one rule fixed in advance: mean validation AUC-ROC over three seeds, a tie band of $0.002$, and ties broken toward the default. On Photo and Computers every OUTPOST arm falls inside the band, so tuning selects nothing, whereas choosing the best arm on the test set would add $0.016$ and $0.004$ (Table~\ref{tab:hparam}). Four selections departed from the default; we re-tested three at ten seeds, and two did not survive (Table~\ref{tab:selsurvive}, Appendix~\ref{app:bonus}). For OUTPOST on Amazon, the validation margin fell from $+0.0057$ to $-0.0004$ and the selected arm proved $0.0040$ worse on test, on nine of ten seeds ($p=0.004$). For NSReg on Photo, the margin fell to $+0.0008$ and the test gain vanished. Only NSReg on Amazon held ($+0.0054$, $p=0.006$), and Table~\ref{tab:main} uses it.

Figure~\ref{fig:noise} (left) and Table~\ref{tab:selnoise} show why. Across 15 to 1{,}076 pairs of arms per graph, the standard error of a paired three-seed difference in validation AUC-ROC is $0.0038$ to $0.0082$, two to four times the tie band. Ten seeds leave it at $0.0021$ to $0.0045$, and reaching $0.002$ would take 11 to 51 seeds. A tie band below the noise of the statistic it reads does not select on quality; the same mechanism selected the weaker SimSample setting on T-Finance (Section~\ref{sec:rq4}). A selection threshold should therefore be set against the paired standard error at the seed count used; under a one-standard-error requirement, every selection in this study would have kept its default.

\subsection{Benchmark Provenance Sets the Size of the Oracle Bonus}\label{sec:rq6}
The size of the effect in Section~\ref{sec:rq1} follows how the anomaly labels were produced. The oracle bonus of OUTPOST is $0.080$ on Computers, $0.079$ on Photo, and $0.045$ on CS; $0.020$ and $0.017$ on the two OGB graphs; and $0.014$, $0.011$, and $0.002$ on T-Finance, Amazon, and Yelp (Fig.~\ref{fig:teaser}, left). Every semi-synthetic graph lies above every OGB graph, and every OGB graph above every graph with observed fraud, with no overlap among the eight means. T-Finance entered the study after this ordering was stated and fell inside the real band. Graph size does not explain the ordering: ogbn-arxiv has $169{,}343$ nodes with relabeled anomalies and a bonus of $0.017$, below all three much smaller semi-synthetic graphs. The bonus measures how far the test curve still moves between epochs late in training, so a benchmark on which training settles leaves the maximum over epochs little to collect.

Against the differences the protocol exists to detect, the bonus is large: the 14 paired differences between methods at the oracle rule in Table~\ref{tab:paired} have a median absolute value of $0.014$, and 13 of them are smaller than $0.045$, the smallest semi-synthetic bonus. On those benchmarks, the choice of selection rule moves a reported score further than the choice of method. Restricting the score to unseen classes widens the bonus to $0.178$ on Photo (Table~\ref{tab:inflation}, Appendix~\ref{app:bonus}).

\begin{table}[t]
  \centering
  \begin{minipage}[t]{0.52\linewidth}
    \centering
    \caption{OUTPOST minus each baseline in AUC-ROC, paired by seed over ten seeds, under both rules. A star marks a Wilcoxon $p$ below 0.05. OUTPOST is significantly ahead of both baselines under both rules on CS, and of DEMO on Computers and Yelp. Win counts and AUC-PR are in Tables~\ref{tab:pdemo} and~\ref{tab:pnsreg}.}
    \label{tab:paired}
    \footnotesize
    \setlength{\tabcolsep}{3pt}
    \begin{tabular}{@{}lcccc@{}}
      \toprule
      & \multicolumn{2}{c}{Best epoch on test} & \multicolumn{2}{c}{Best epoch on valid.} \\
      \cmidrule(lr){2-3}\cmidrule(lr){4-5}
      Graph & $-$\,DEMO & $-$\,NSReg & $-$\,DEMO & $-$\,NSReg \\
      \midrule
      Photo & $-0.018$ & $-0.006$ & $+0.017$ & $-0.002$ \\
      Computers & $+0.073^{*}$ & $+0.014^{*}$ & $+0.044^{*}$ & $-0.014$ \\
      CS & $+0.020^{*}$ & $+0.014^{*}$ & $+0.038^{*}$ & $+0.010^{*}$ \\
      Yelp & $+0.014^{*}$ & $+0.006$ & $+0.018^{*}$ & $+0.018^{*}$ \\
      Amazon & $-0.002$ & $-0.005$ & $-0.009^{*}$ & $-0.012^{*}$ \\
      T-Finance & $-0.007$ & $-0.015^{*}$ & $-0.008$ & $-0.021^{*}$ \\
      ogbn-arxiv & $+0.005$ & $-0.030^{*}$ & $+0.002$ & $-0.024^{*}$ \\
      \bottomrule
    \end{tabular}
  \end{minipage}\hfill
  \begin{minipage}[t]{0.44\linewidth}
    \centering
    \caption{Change in AUC-ROC when one component of OUTPOST is removed, at the best epoch on test and the development budget (200 epochs on semi-synthetic graphs, 400 elsewhere). A star marks $p<0.05$; a dagger marks a unanimous result at five seeds, where the smallest attainable $p$ is 0.0625. Pseudo-labeling carries the semi-synthetic gains, and removing it improves Yelp and T-Finance.}
    \label{tab:decomp}
    \footnotesize
    \setlength{\tabcolsep}{3pt}
    \begin{tabular}{@{}lccc@{}}
      \toprule
      Graph & Gate & Conformal & Pseudo-lab. \\
      \midrule
      \multicolumn{4}{@{}l}{\emph{Semi-synthetic}} \\
      Photo & $+0.000$ & $-0.004$ & $-0.038^{\dagger}$ \\
      Computers & $-0.001$ & $-0.037^{\dagger}$ & $-0.065^{\dagger}$ \\
      CS & $+0.000$ & $-0.005^{\dagger}$ & $-0.065^{\dagger}$ \\
      \addlinespace[2pt]
      \multicolumn{4}{@{}l}{\emph{Real}} \\
      Yelp & $+0.001$ & $-0.000$ & $+0.006^{*}$ \\
      Amazon & $-0.000$ & $+0.001$ & $-0.002$ \\
      T-Finance & $-0.000$ & $+0.003$ & $+0.010^{*}$ \\
      \bottomrule
    \end{tabular}
  \end{minipage}
\end{table}

This matters for ablations. At 200 epochs, removing pseudo-labeling and the atlas gate together lowers the oracle AUC-ROC of OUTPOST on Photo by $0.038$ on ten of ten seeds, but costs only $0.015$ under the validation rule, not significantly ($p=0.19$): $62\%$ of the apparent benefit is variance that only the oracle rule collects, and the share is $43\%$ on Computers and $44\%$ on CS (Table~\ref{tab:variance}). Judged by the oracle rule alone, a component can look useful partly because it destabilizes training.

The provenance split also appears in graph structure. A training-free diagnostic, prototypes fitted on $5\%$ of the normal nodes and applied to features propagated by zero to three hops, assigns each of 37 anomaly classes an AUC-ROC that correlates with the class's median same-class neighbor fraction (Spearman $\rho=0.778$, within-graph permutation $p=5\times10^{-5}$), consistent with the low-pass reading of message passing \citep{nt2019lowpass,oono2020oversmoothing,zhu2020beyond}. The relation describes these graphs rather than predicting new ones (Appendices~\ref{app:diag} and~\ref{app:theory}), but one implication holds: Photo, Computers, and CS sit at the homophilous end of the range, at $0.55$ to $1.00$, so a method tuned on them is tuned for clustered anomalies.

\subsection{Component Contributions Depend on Anomaly Provenance}\label{sec:rq4}\label{sec:rq5}
Removing one component at a time, paired by seed (Table~\ref{tab:decomp}, Fig.~\ref{fig:teaser} right; full set in Table~\ref{tab:ablfull}), shows that the parts of Section~\ref{sec:method} do not contribute equally, and that which parts matter depends on provenance. Pseudo-labeling carries the gains on the semi-synthetic graphs: removing it costs $0.038$ to $0.065$ on Photo, Computers, and CS, on every seed. On the graphs with observed fraud it costs nothing: removal is neutral on Amazon ($-0.0018$, not significant) and improves Yelp ($+0.0059$, $p=0.027$) and T-Finance ($+0.0097$, $p=0.004$; AUC-PR $+0.035$). Conformal calibration of the pseudo-label threshold matters on Computers, where fixing the threshold at DEMO's $0.95$ costs $0.037$, and little elsewhere. The atlas gate has no measurable effect at the main protocol, moving AUC-ROC by at most $0.0016$ across seven graphs (Table~\ref{tab:gateoff}); we keep it because the reported numbers include it.

T-Finance rules out the natural explanation, that real fraud is scattered among normal nodes so that no propagation-based signal can help. Its fraud nodes have a median same-class neighbor fraction of $0.629$, inside the semi-synthetic range of $0.55$ to $1.00$ and four times that of Yelp ($0.160$), yet pseudo-labeling hurts there as well, as we registered before training on T-Finance (P28). We attribute the difference to how the labels were constructed rather than to topology: a relabeled class is a coherent cluster by construction, so confident predictions on unlabeled nodes are mostly correct and self-training reinforces them, whereas observed fraud offers no such guarantee. An ablation confined to relabeled-class graphs therefore does not show that a component helps on real anomalies.

SimSample is the one component whose effect admits a causal test, and on Yelp it is the only component whose removal measurably lowers the score: disabling it costs $0.0504$ AUC-PR and $0.0263$ AUC-ROC on ten of ten seeds. Three tests separate the similarity order from the sampling determinism it also introduces (Fig.~\ref{fig:noise}, right; Appendix~\ref{app:simsample}). A placebo that keeps the deterministic sampling but shuffles the order reaches $0.3269$ AUC-PR, below the $0.3320$ of the disabled arm, so the gain is not an artifact of reduced sampling randomness. A dose--response sweep raises AUC-PR monotonically with the similarity fraction, from $0.3320$ to $0.3525$, $0.3632$, $0.3696$, and $0.3824$. Raising the first-hop budget from 25 to 50, 100, and 200 shrinks the gain from $0.0504$ to $0.0394$, $0.0269$, and $0.0217$ (Spearman $\rho=-0.70$ over 25 paired runs, $p<10^{-4}$), mostly because the enabled arm falls, by $0.020$; the disabled arm rises by $0.010$ at the first step and is flat thereafter. The gain therefore comes from which neighbors enter the sample.

Across graphs, SimSample helps Yelp and T-Finance, hurts Amazon and Photo, and is neutral on CS and ogbn-arxiv; our registered account in terms of node degree and sampling budget does not explain this pattern (P13, P14; Appendix~\ref{app:simsample}). On T-Finance the validation rule selected the weaker arm; Table~\ref{tab:main} reports that arm as registered, which understates OUTPOST there by $0.0128$ AUC-ROC.

\section{Conclusion}\label{sec:conclusion}

We audited open-set GAD under one pre-registered protocol, re-running DEMO and NSReg alongside OUTPOST, a first-order detector built for this study, and scoring every run at the epoch a practitioner could choose as well as at the most favorable one. A comparison in this setting is not settled by its methods alone: which detector leads depends on the rule that selects the reported epoch, and which component helps depends on whether the anomalies were observed or constructed. On the semi-synthetic benchmarks this literature uses most, the selection rule moves a score further than the choice of detector, and transcribed baselines can reorder a table. These are findings about measurement, not method quality: 12 of the 14 oracle-rule differences between OUTPOST and the two baselines fall at or below $0.020$ AUC-ROC, and OUTPOST leads on CS and Yelp under both rules and alone scales to ogbn-mag. A reported gain in open-set GAD should therefore state the rule that selected it and the source of its anomalies. We close with seven reporting practices that follow from these results (Appendix~\ref{app:checklist}), chief among them reporting both selection rules, re-running baselines under one protocol, and ablating on observed anomalies.

\textbf{Limitations and future work.} Our conclusions are conditioned on GraphSAGE-type low-pass encoders. All methods share one split generator and one label budget, and a different budget could reorder small differences. The provenance distinction rests on three graphs with observed labels, too few to separate provenance from other properties they share, and all three are binary, so they test real anomalies but not unseen classes. The oracle bonus is measured on our runs of three methods; for other published methods, it is the bonus available under their protocol, not a measured one. The baseline tuning grids exclude CS for cost, and the DEMO selection on Amazon was not retested. Several ablation cells have five seeds and cannot reach $p<0.05$; we read them from unanimity and effect size. These limitations set the next step: extending the audit to other encoder families and to observed anomalies with several classes.

\bibliography{refs}
\bibliographystyle{iclr2027_conference}

\appendix
\raggedbottom
\setlength{\textfloatsep}{10pt plus 2pt minus 2pt}
\setlength{\floatsep}{10pt plus 2pt minus 2pt}
\setlength{\intextsep}{10pt plus 2pt minus 2pt}
\section*{Overview of the Appendix}
Appendix~\ref{app:data} describes the graphs. Appendix~\ref{app:hparams} lists settings, hardware, and sensitivity. Appendix~\ref{app:baselines} describes the baseline configurations. Appendix~\ref{app:published} gives the full published field and the AUC-PR version of the main table. Appendix~\ref{app:paired} contains all paired tests, the unseen-class results, and the rankings. Appendix~\ref{app:bonus} covers the oracle bonus, the training budget, and tuning. Appendix~\ref{app:ablations} holds all ablations. Appendix~\ref{app:simsample} covers SimSample. Appendix~\ref{app:diag} gives the training-free diagnostic, and Appendix~\ref{app:theory} a signal-to-noise analysis of propagation. Appendix~\ref{app:efficiency} reports model size and memory. Appendix~\ref{app:checklist} gives the reporting checklist, and Appendix~\ref{app:prereg} the pre-registration ledger.

\section{Graphs}\label{app:data}

Photo, Computers, and CS are co-purchase and co-authorship graphs \citep{shchur2018pitfalls}. For ogbn-mag we use the paper nodes and citation edges \citep{hu2020ogb}; its anomaly classes are venues with 129 to 199 papers among 736{,}389, and for 11 of 15 classes the median anomaly has no same-class neighbor. Yelp, Amazon, and T-Finance are used as released by earlier GAD work \citep{rayana2015yelp,mcauley2013amazon,tang2022bwgnn,dou2020caregnn}, with all relations merged into one edge set. The same-class fraction shows that ``real'' does not mean ``scattered'': Yelp (0.16) and Amazon (0.08) are scattered, while T-Finance (0.63) is as homophilous as the semi-synthetic graphs. Together with the ogbn-mag classes that have no same-class neighbor, this contradicts the registered prediction that real fraud is more scattered than every relabeled class (P4).

\begin{table}[H]
  \centering
  \caption{The eight graphs, grouped by how their anomaly labels were produced. Same-class fraction is the median share of a class node's neighbors with the same label, given as a range over the graph's anomaly classes; it uses true labels and serves only as a diagnostic. The median degree of ogbn-mag was not recorded.}
  \label{tab:datasets}
  \small
  \begin{tabular}{@{}lrrrrcl@{}}
    \toprule
    Graph & Nodes & Features & Median deg. & Rotations & Unseen classes & Same-class frac. \\
    \midrule
    \multicolumn{7}{@{}c}{\emph{Semi-synthetic: classes holding at most 5\% of the nodes relabeled as anomalous}} \\
    Photo & 7{,}650 & 745 & 22 & 2 & yes & 0.56\textendash{}0.94 \\
    Computers & 13{,}752 & 767 & 22 & 5 & yes & 0.55\textendash{}0.94 \\
    CS & 18{,}333 & 6{,}805 & 6 & 8 & yes & 0.64\textendash{}1.00 \\
    \addlinespace[2pt]
    \multicolumn{7}{@{}c}{\emph{OGB: classes relabeled within a size band, 3\textendash{}5\% for ogbn-arxiv and at most 0.03\% for ogbn-mag}} \\
    ogbn-arxiv & 169{,}343 & 128 & 1 & 4 & yes & 0.46\textendash{}0.82 \\
    ogbn-mag & 736{,}389 & 128 & -- & 15 & yes & 0.00\textendash{}1.00 \\
    \addlinespace[2pt]
    \multicolumn{7}{@{}c}{\emph{Real: anomaly labels are observed fraud}} \\
    Yelp & 45{,}954 & 32 & 168 & 1 & no & 0.16 \\
    Amazon & 11{,}944 & 25 & 403 & 1 & no & 0.08 \\
    T-Finance & 39{,}357 & 10 & 265 & 1 & no & 0.63 \\
    \bottomrule
  \end{tabular}
\end{table}

\section{Settings, Hardware, and Sensitivity}\label{app:hparams}

\textbf{OUTPOST.} Adam \citep{kingma2015adam}, learning rate $10^{-3}$, weight decay $5\times10^{-4}$, two GraphSAGE layers, fan-out $(25,10)$, batch size 512 (256 on ogbn-mag), one optimizer step per epoch. \emph{Semi-synthetic and OGB graphs:} hidden width $h=64$, dropout 0.5, $K=8$ prototypes, warm-up $W=5$ epochs (10 on OGB), mixup, halo, and atlas loss on, $\rho=0$. \emph{Real graphs:} $h=32$, dropout 0.2, $K=3$, $W=10$, no synthesis and no atlas loss; $\rho=1$ on Yelp, and chosen on validation from $\{0,1\}$ on Amazon and T-Finance (both chose 0). Shared values: $\lambda_{\text{un}}=1.0$, $\lambda_{\text{mix}}=\lambda_{\text{halo}}=0.2$, $\lambda_{\text{atlas}}=0.5$, margin $m=0.1$, $\alpha_+=0.05$, $\tau_-=0.05$, $\alpha_r=0.1$, moving-average rate 0.05, gate quantile 0.9, $\eta\sim\mathcal{U}[1.2,2.0]$. The weak view adds Gaussian noise ($\sigma=0.02$) and masks 10\% of features; the strong view also mixes features ($\alpha=0.1$) and rescales them ($\gamma=0.1$). Amazon and T-Finance inherit the Yelp block unchanged except for the input width.

\textbf{Hardware and software.} NVIDIA RTX A6000 GPUs, Python 3.10.12, PyTorch 2.7.1 \citep{paszke2019pytorch} with CUDA 12.6, PyTorch Geometric 2.7.0 \citep{fey2019pyg}. Deterministic algorithms are enabled. Re-running one Amazon configuration on a second machine with an RTX A5000 and the same software gave metrics identical to six decimals. Sharding rotations across GPUs and holding the feature table on the GPU were each tested and gave identical per-rotation metrics. We make no claim about relative training speed: our timings were taken on shared machines, and the same Photo pair gave time ratios of 0.75 and 1.40 on two occasions. The one timing we report, the 6.5$\times$ cost of DEMO's energy term (Appendix~\ref{app:baselines}), lies far outside that range.

\textbf{Sensitivity.} Five seeds per swept value against the ten-seed default. For each hyperparameter, we compare the range of the sweep with the seed standard deviation (sd) of the default. On Yelp, hidden width 16/32/64 gives a range of 0.0078 AUC-ROC (0.5$\times$ the seed sd) and 0.0350 AUC-PR (1.03$\times$); $\alpha_+\in\{0.01,0.05,0.10\}$ gives 0.0013 and 0.0047 (0.09$\times$ and 0.14$\times$). On Photo, $\lambda_{\text{un}}\in\{0.5,1.0,1.25,1.5\}$ gives 0.0116 and 0.0107 (0.39$\times$ and 0.48$\times$). Five of six sweeps are well inside the seed noise.

\section{Baseline Configurations}\label{app:baselines}

\textbf{DEMO.} We follow the published configuration, including multi-sample mixup over a dense personalized-PageRank (PPR) matrix built with the authors' recipe. The released entry script disables mixup by default, which corresponds to the ``w/o Mix'' ablation of \citet{yu2026demo}; all our DEMO runs enable it. DEMO's energy-gradient reweighting needs a second-order backward pass per labeled node per epoch. On Photo, at three matched seeds, it gives 0.8408 AUC-ROC against 0.8395 without it ($p=0.83$) and 0.5236 against 0.5235 AUC-PR ($p=0.99$), at 6.5$\times$ the compute (2{,}196 against 337 seconds per rotation), so we omit it on all graphs. Three changes to the code were needed. We pass CPU indices to a sampler that received GPU indices for a CPU tensor and failed before the first epoch. We train for 400 epochs instead of the configured 200, as for every method, which can only help DEMO under the oracle rule. We build the evaluation loader once instead of every epoch, which changes no metric because the sampler draws no random numbers at construction.

\textbf{NSReg.} A wrapper feeds our tensors and splits to the released trainer, whose source we do not modify. We apply the single released configuration to every graph, changing only the input width, and tune NSReg with the same validation-only rule as OUTPOST (Table~\ref{tab:bsel}). Three environment changes were needed: a replacement for a sampler that depends on an unavailable library, a type conversion for index arrays, and chunking of the full-graph pass on CS to fit in memory.

\textbf{ogbn-mag.} Neither baseline can run on ogbn-mag. DEMO is defined on a dense $N\times N$ PPR matrix, which takes 2.17\,TB at $N=736{,}389$, against 115\,GB on ogbn-arxiv. NSReg's regularizer ranges over all ordered pairs of labeled normal nodes: 1.34 billion pairs on ogbn-mag against 50 million on ogbn-arxiv, and a probe used 280\,GB of host memory without finishing setup. Subsampling the pairs would run, but would no longer be NSReg. Both baselines run on ogbn-arxiv.

\textbf{Re-runs against published values.} Table~\ref{tab:repro} compares our re-runs of the released code with the values published for DEMO and NSReg.

\begin{table}[H]
  \centering
  \caption{Published AUC-ROC against our re-runs of the released code, scored at the best epoch on test, at each method's released budget and at the 400 epochs used for every method in this study. At 400 epochs, five of seven re-runs exceed the published value: NSReg on all four graphs and DEMO on CS. Most of DEMO's Photo gap closes with the longer budget.}
  \label{tab:repro}
  \small
  \begin{tabular}{@{}lccccc@{}}
    \toprule
    & & & \multicolumn{2}{c}{Re-run by us} & \\
    \cmidrule(lr){4-5}
    Method & Graph & Published & Released budget & 400 epochs & Difference \\
    \midrule
    \multicolumn{6}{@{}c}{\emph{DEMO, released budget 200 epochs}} \\
    DEMO & Photo & 0.9023 & 0.8403 & 0.8879 & $-0.0144$ \\
    DEMO & Computers & 0.8439 & 0.7685 & 0.7782 & $-0.0657$ \\
    DEMO & CS & 0.9448 & 0.9603 & 0.9646 & $+0.0198$ \\
    \addlinespace[2pt]
    \multicolumn{6}{@{}c}{\emph{NSReg, released budget 201 epochs}} \\
    NSReg & Photo & 0.8360 & 0.8121 & 0.8760 & $+0.0400$ \\
    NSReg & Computers & 0.7403 & 0.8085 & 0.8370 & $+0.0967$ \\
    NSReg & CS & 0.9032 & 0.9709 & 0.9701 & $+0.0669$ \\
    NSReg & Yelp & 0.7015 & 0.7085 & 0.7388 & $+0.0373$ \\
    \bottomrule
  \end{tabular}
\end{table}

\section{Published Field and AUC-PR}\label{app:published}

\begin{figure}[H]
  \centering
  \includegraphics[width=0.92\linewidth]{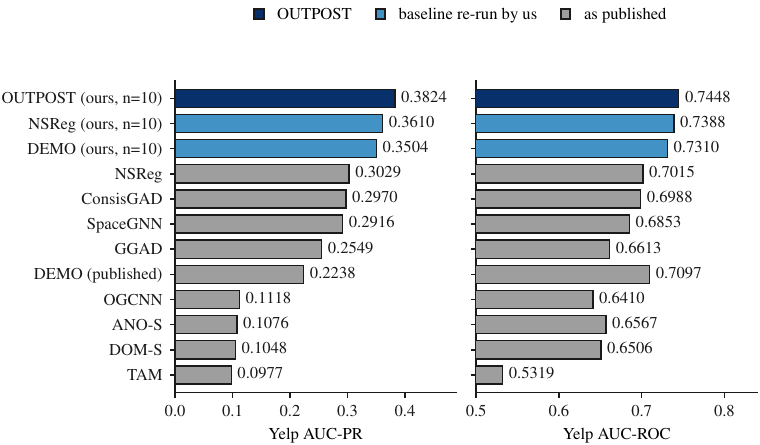}
  \caption{Yelp, the one graph in this study that carries a published field and whose anomalies were observed rather than relabeled. Gray bars are published values as reported by \citet{yu2026demo}; the two blue shades are our runs under the protocol of Section~\ref{sec:protocol}. OUTPOST reaches 0.3824 AUC-PR and leads the re-run NSReg (0.3610) and DEMO (0.3504). Most of its margin over the best published value (0.3029) comes from the change of protocol, which lifts the re-run baselines as well. Methods are ordered by AUC-PR, so the AUC-ROC panel is not monotone.}
  \label{fig:yelp}
\end{figure}

Tables~\ref{tab:pubsmall} and~\ref{tab:publarge} list all values transcribed from \citet{yu2026demo}, grouped as in that source: unsupervised detectors \citep{peng2018anomalous,ding2019dominant,fan2020anomalydae,chen2020gaan,liu2021cola,xu2022conad}, and semi-supervised and open-set detectors \citep{chen2024consisgad,qiao2024ggad,qiao2023tam,wang2021ocgnn,dong2025spacegnn,wang2025nsreg,bendale2016openmax,yu2026demo}, including ANO-S and DOM-S, the semi-supervised variants of ANOMALOUS and DOMINANT. Transcribed rows come from other splits, budgets, and seeds; the three rows below the rule are ours. Figure~\ref{fig:yelp} shows the Yelp column, and Table~\ref{tab:mainpr} gives the AUC-PR counterpart of Table~\ref{tab:main}.

\begin{table}[H]
  \centering
  \caption{The published field on the three semi-synthetic graphs, transcribed from \citet{yu2026demo}, with our three re-run rows below the rule. Transcribed rows come from other splits, budgets, and seed counts, so the two blocks are not comparable with each other; the best and runner-up are therefore marked within each block separately, in bold and underlined.}
  \label{tab:pubsmall}
  \small
  \begin{tabular}{@{}lcccccc@{}}
    \toprule
    & \multicolumn{2}{c}{Photo} & \multicolumn{2}{c}{Computers} & \multicolumn{2}{c}{CS} \\
    \cmidrule(lr){2-3}\cmidrule(lr){4-5}\cmidrule(lr){6-7}
    Method & AUC-ROC & AUC-PR & AUC-ROC & AUC-PR & AUC-ROC & AUC-PR \\
    \midrule
    \multicolumn{7}{@{}c}{\emph{Unsupervised}} \\
    ANOMALOUS & 0.5574 & 0.0879 & 0.5737 & 0.1693 & 0.2997 & 0.1634 \\
    DOMINANT & 0.4716 & 0.0837 & 0.5450 & 0.1644 & 0.4029 & 0.1886 \\
    AnomalyDAE & 0.4179 & 0.0770 & 0.5658 & 0.1723 & 0.3978 & 0.1864 \\
    GAAN & 0.4346 & 0.0710 & 0.5595 & 0.1796 & 0.4646 & 0.2111 \\
    CoLA & 0.5618 & 0.0989 & 0.4897 & 0.1472 & 0.4353 & 0.2029 \\
    CONAD & 0.4763 & 0.0862 & 0.5445 & 0.1619 & 0.4028 & 0.1886 \\
    \addlinespace[2pt]
    \multicolumn{7}{@{}c}{\emph{Semi-supervised and open-set}} \\
    ConsisGAD & \underline{0.8668} & \underline{0.5987} & 0.6250 & 0.3572 & 0.7178 & 0.5271 \\
    GGAD & 0.7976 & 0.5677 & 0.7210 & 0.4529 & \underline{0.9081} & \underline{0.8198} \\
    TAM & 0.6045 & 0.1084 & 0.4432 & 0.1355 & 0.6398 & 0.3542 \\
    OCGNN & 0.6279 & 0.1323 & 0.5049 & 0.1505 & 0.7819 & 0.4926 \\
    ANO-S & 0.5730 & 0.1097 & 0.4628 & 0.1392 & 0.8380 & 0.6401 \\
    DOM-S & 0.5785 & 0.1107 & 0.4488 & 0.1330 & 0.8445 & 0.6382 \\
    SpaceGNN & 0.8030 & 0.5271 & \underline{0.8296} & \underline{0.6439} & 0.7784 & 0.6587 \\
    NSReg & 0.8360 & 0.4777 & 0.7403 & 0.5437 & 0.9032 & 0.8115 \\
    GNN+OpenMax & 0.7618 & 0.3942 & 0.6713 & 0.3942 & 0.8213 & 0.7559 \\
    DEMO & \textbf{0.9023} & \textbf{0.6330} & \textbf{0.8439} & \textbf{0.6458} & \textbf{0.9448} & \textbf{0.8857} \\
    \midrule
    \multicolumn{7}{@{}c}{\emph{Re-run by us under one protocol, ten seeds}} \\
    DEMO & \textbf{0.8879} & \underline{0.5848} & 0.7782 & 0.5289 & 0.9646 & 0.9155 \\
    NSReg & \underline{0.8760} & \textbf{0.5902} & \underline{0.8370} & \underline{0.6055} & \underline{0.9701} & \underline{0.9336} \\
    OUTPOST & 0.8703 & 0.5557 & \textbf{0.8510} & \textbf{0.6521} & \textbf{0.9842} & \textbf{0.9575} \\
    \bottomrule
  \end{tabular}
\end{table}

\begin{table}[H]
  \centering
  \caption{The published field on the three large graphs, transcribed from \citet{yu2026demo}, with our three re-run rows below the rule. As in Table~\ref{tab:pubsmall}, the best and runner-up are marked within each block separately, because transcribed and re-run values are not comparable with each other. Amazon and T-Finance carry no published open-set values and are omitted. A dash is a value the source does not report, or a run that cannot be made on ogbn-mag (Appendix~\ref{app:baselines}).}
  \label{tab:publarge}
  \small
  \begin{tabular}{@{}lcccccc@{}}
    \toprule
    & \multicolumn{2}{c}{Yelp} & \multicolumn{2}{c}{ogbn-arxiv} & \multicolumn{2}{c}{ogbn-mag} \\
    \cmidrule(lr){2-3}\cmidrule(lr){4-5}\cmidrule(lr){6-7}
    Method & AUC-ROC & AUC-PR & AUC-ROC & AUC-PR & AUC-ROC & AUC-PR \\
    \midrule
    \multicolumn{7}{@{}c}{\emph{As published}} \\
    ConsisGAD & 0.6988 & \underline{0.2970} & \underline{0.6216} & 0.3148 & \underline{0.4909} & \underline{0.0043} \\
    GGAD & 0.6613 & 0.2549 & 0.6007 & 0.2843 & -- & -- \\
    TAM & 0.5319 & 0.0977 & -- & -- & -- & -- \\
    OCGNN & 0.6410 & 0.1118 & -- & -- & -- & -- \\
    ANO-S & 0.6567 & 0.1076 & 0.4510 & 0.1463 & -- & -- \\
    DOM-S & 0.6506 & 0.1048 & 0.4505 & 0.1482 & -- & -- \\
    SpaceGNN & 0.6853 & 0.2916 & 0.6133 & \underline{0.3301} & 0.4626 & \underline{0.0043} \\
    NSReg & \underline{0.7015} & \textbf{0.3029} & 0.6182 & 0.3230 & 0.4836 & 0.0041 \\
    DEMO & \textbf{0.7097} & 0.2238 & \textbf{0.6364} & \textbf{0.3329} & \textbf{0.4967} & \textbf{0.0054} \\
    \midrule
    \multicolumn{7}{@{}c}{\emph{Re-run by us under one protocol, ten seeds}} \\
    DEMO & 0.7310 & 0.3504 & 0.6176 & 0.2966 & -- & -- \\
    NSReg & \underline{0.7388} & \underline{0.3610} & \textbf{0.6533} & \textbf{0.3110} & -- & -- \\
    OUTPOST & \textbf{0.7448} & \textbf{0.3824} & \underline{0.6229} & \underline{0.3049} & 0.5928 & 0.0101 \\
    \bottomrule
  \end{tabular}
\end{table}

\begin{table}[H]
  \centering
  \caption{AUC-PR under the same protocol as Table~\ref{tab:main}, as mean and standard deviation over ten seeds. On each graph and under each rule, the best of the three methods is bold and the runner-up underlined. The published column, in gray, is the highest transcribed value; it is DEMO's on every graph except Yelp, where it is NSReg's.}
  \label{tab:mainpr}
  \small
  \setlength{\tabcolsep}{5pt}
  \begin{tabular}{@{}lccccccc@{}}
    \toprule
    & \pub{Best} & \multicolumn{3}{c}{Best epoch on test} & \multicolumn{3}{c}{Best epoch on validation} \\
    \cmidrule(lr){3-5}\cmidrule(lr){6-8}
    Graph & \pub{published} & DEMO & NSReg & OUTPOST & DEMO & NSReg & OUTPOST \\
    \midrule
    \multicolumn{8}{@{}c}{\emph{Semi-synthetic}} \\
    Photo & \pub{0.633} & \underline{0.585}\,{\scriptsize$\pm$.020} & \textbf{0.590}\,{\scriptsize$\pm$.059} & 0.556\,{\scriptsize$\pm$.036} & 0.487\,{\scriptsize$\pm$.042} & \textbf{0.506}\,{\scriptsize$\pm$.058} & \underline{0.498}\,{\scriptsize$\pm$.037} \\
    Computers & \pub{0.646} & 0.529\,{\scriptsize$\pm$.025} & \underline{0.606}\,{\scriptsize$\pm$.020} & \textbf{0.652}\,{\scriptsize$\pm$.009} & 0.458\,{\scriptsize$\pm$.029} & \underline{0.526}\,{\scriptsize$\pm$.033} & \textbf{0.532}\,{\scriptsize$\pm$.034} \\
    CS & \pub{0.886} & 0.915\,{\scriptsize$\pm$.011} & \underline{0.934}\,{\scriptsize$\pm$.007} & \textbf{0.958}\,{\scriptsize$\pm$.004} & 0.795\,{\scriptsize$\pm$.036} & \underline{0.855}\,{\scriptsize$\pm$.034} & \textbf{0.868}\,{\scriptsize$\pm$.025} \\
    \addlinespace[2pt]
    \multicolumn{8}{@{}c}{\emph{OGB}} \\
    ogbn-arxiv & \pub{0.333} & 0.297\,{\scriptsize$\pm$.007} & \textbf{0.311}\,{\scriptsize$\pm$.005} & \underline{0.305}\,{\scriptsize$\pm$.006} & 0.287\,{\scriptsize$\pm$.007} & \textbf{0.300}\,{\scriptsize$\pm$.010} & \underline{0.295}\,{\scriptsize$\pm$.008} \\
    ogbn-mag & \pub{0.005} & -- & -- & 0.010\,{\scriptsize$\pm$.000} & -- & -- & 0.009\,{\scriptsize$\pm$.001} \\
    \addlinespace[2pt]
    \multicolumn{8}{@{}c}{\emph{Real}} \\
    Yelp & \pub{0.303} & 0.350\,{\scriptsize$\pm$.022} & \underline{0.361}\,{\scriptsize$\pm$.020} & \textbf{0.382}\,{\scriptsize$\pm$.034} & 0.337\,{\scriptsize$\pm$.023} & \underline{0.338}\,{\scriptsize$\pm$.025} & \textbf{0.371}\,{\scriptsize$\pm$.034} \\
    Amazon & \pub{--} & \underline{0.839}\,{\scriptsize$\pm$.011} & \textbf{0.841}\,{\scriptsize$\pm$.012} & \underline{0.839}\,{\scriptsize$\pm$.009} & \textbf{0.830}\,{\scriptsize$\pm$.013} & \underline{0.829}\,{\scriptsize$\pm$.017} & 0.818\,{\scriptsize$\pm$.022} \\
    T-Finance & \pub{--} & \underline{0.712}\,{\scriptsize$\pm$.026} & \textbf{0.726}\,{\scriptsize$\pm$.034} & 0.681\,{\scriptsize$\pm$.040} & \underline{0.656}\,{\scriptsize$\pm$.074} & \textbf{0.668}\,{\scriptsize$\pm$.045} & 0.643\,{\scriptsize$\pm$.038} \\
    \bottomrule
  \end{tabular}
\end{table}

\section{Paired Tests, Unseen Classes, and Rankings}\label{app:paired}

Tables~\ref{tab:pdemo} and~\ref{tab:pnsreg} give every paired comparison between OUTPOST and each baseline, with win counts and both metrics; Table~\ref{tab:unseen} repeats the comparison on anomalies of classes never seen in training. Tables~\ref{tab:rankall} and~\ref{tab:rankunseen} give the resulting order of the three methods under each rule, over all anomalies and over unseen classes only.

\begin{table}[H]
  \centering
  \caption{OUTPOST minus DEMO on every graph both methods run, paired by seed over ten seeds and scored under both rules. Wins are seeds on which OUTPOST is ahead; there were no ties. At ten paired seeds, the smallest attainable Wilcoxon $p$ is 0.002.}
  \label{tab:pdemo}
  \small
  \begin{tabular}{@{}lrrrrrrrr@{}}
    \toprule
    & \multicolumn{4}{c}{Best epoch on test} & \multicolumn{4}{c}{Best epoch on validation} \\
    \cmidrule(lr){2-5}\cmidrule(lr){6-9}
    & \multicolumn{2}{c}{AUC-ROC} & \multicolumn{2}{c}{AUC-PR} & \multicolumn{2}{c}{AUC-ROC} & \multicolumn{2}{c}{AUC-PR} \\
    \cmidrule(lr){2-3}\cmidrule(lr){4-5}\cmidrule(lr){6-7}\cmidrule(lr){8-9}
    Graph & $\Delta$ & Wins, $p$ & $\Delta$ & Wins, $p$ & $\Delta$ & Wins, $p$ & $\Delta$ & Wins, $p$ \\
    \midrule
    Photo & $-0.0176$ & 3, .065 & $-0.0291$ & 1, .037 & $+0.0170$ & 7, .193 & $+0.0107$ & 8, .105 \\
    Computers & $+0.0728$ & 10, .002 & $+0.1232$ & 10, .002 & $+0.0441$ & 10, .002 & $+0.0747$ & 10, .002 \\
    CS & $+0.0196$ & 10, .002 & $+0.0419$ & 10, .002 & $+0.0381$ & 10, .002 & $+0.0732$ & 10, .002 \\
    Yelp & $+0.0138$ & 9, .019 & $+0.0320$ & 10, .002 & $+0.0180$ & 9, .027 & $+0.0334$ & 9, .004 \\
    Amazon & $-0.0021$ & 3, .193 & $+0.0002$ & 5, .922 & $-0.0085$ & 0, .002 & $-0.0116$ & 2, .105 \\
    T-Finance & $-0.0074$ & 4, .193 & $-0.0315$ & 3, .084 & $-0.0083$ & 3, .160 & $-0.0128$ & 3, .492 \\
    ogbn-arxiv & $+0.0053$ & 8, .160 & $+0.0083$ & 10, .002 & $+0.0022$ & 7, .105 & $+0.0077$ & 10, .002 \\
    \bottomrule
  \end{tabular}
\end{table}

\begin{table}[H]
  \centering
  \caption{OUTPOST minus NSReg, on the same seeds and in the same form as Table~\ref{tab:pdemo}. NSReg is the stronger of the two baselines under the validation rule, and CS and Yelp are the only graphs on which OUTPOST leads it under both rules.}
  \label{tab:pnsreg}
  \small
  \begin{tabular}{@{}lrrrrrrrr@{}}
    \toprule
    & \multicolumn{4}{c}{Best epoch on test} & \multicolumn{4}{c}{Best epoch on validation} \\
    \cmidrule(lr){2-5}\cmidrule(lr){6-9}
    & \multicolumn{2}{c}{AUC-ROC} & \multicolumn{2}{c}{AUC-PR} & \multicolumn{2}{c}{AUC-ROC} & \multicolumn{2}{c}{AUC-PR} \\
    \cmidrule(lr){2-3}\cmidrule(lr){4-5}\cmidrule(lr){6-7}\cmidrule(lr){8-9}
    Graph & $\Delta$ & Wins, $p$ & $\Delta$ & Wins, $p$ & $\Delta$ & Wins, $p$ & $\Delta$ & Wins, $p$ \\
    \midrule
    Photo & $-0.0057$ & 5, .769 & $-0.0346$ & 5, .232 & $-0.0015$ & 6, .922 & $-0.0081$ & 5, .846 \\
    Computers & $+0.0140$ & 9, .014 & $+0.0466$ & 10, .002 & $-0.0140$ & 2, .065 & $+0.0065$ & 6, .492 \\
    CS & $+0.0141$ & 10, .002 & $+0.0238$ & 10, .002 & $+0.0095$ & 8, .027 & $+0.0131$ & 6, .232 \\
    Yelp & $+0.0060$ & 8, .105 & $+0.0214$ & 8, .019 & $+0.0178$ & 9, .027 & $+0.0328$ & 8, .019 \\
    Amazon & $-0.0052$ & 3, .084 & $-0.0018$ & 4, .557 & $-0.0120$ & 2, .027 & $-0.0108$ & 4, .232 \\
    T-Finance & $-0.0151$ & 2, .014 & $-0.0450$ & 3, .065 & $-0.0211$ & 1, .004 & $-0.0250$ & 4, .557 \\
    ogbn-arxiv & $-0.0304$ & 0, .002 & $-0.0061$ & 0, .002 & $-0.0239$ & 0, .002 & $-0.0052$ & 1, .006 \\
    \bottomrule
  \end{tabular}
\end{table}

\begin{table}[H]
  \centering
  \caption{The same paired comparison restricted to anomalies of classes never seen in training. The three real graphs are binary and have no unseen class, so they do not appear. Wins are out of ten seeds.}
  \label{tab:unseen}
  \small
  \setlength{\tabcolsep}{3.5pt}
  \begin{tabular}{@{}llrrrrrrrr@{}}
    \toprule
    & & \multicolumn{4}{c}{Best epoch on test} & \multicolumn{4}{c}{Best epoch on validation} \\
    \cmidrule(lr){3-6}\cmidrule(lr){7-10}
    & & \multicolumn{2}{c}{AUC-ROC} & \multicolumn{2}{c}{AUC-PR} & \multicolumn{2}{c}{AUC-ROC} & \multicolumn{2}{c}{AUC-PR} \\
    \cmidrule(lr){3-4}\cmidrule(lr){5-6}\cmidrule(lr){7-8}\cmidrule(lr){9-10}
    Graph & Against & $\Delta$ & Wins, $p$ & $\Delta$ & Wins, $p$ & $\Delta$ & Wins, $p$ & $\Delta$ & Wins, $p$ \\
    \midrule
    Photo & DEMO & $-0.0126$ & 3, .232 & $-0.0035$ & 4, .625 & $+0.0291$ & 7, .193 & $+0.0031$ & 6, .275 \\
    & NSReg & $-0.0044$ & 5, .625 & $-0.0496$ & 3, .105 & $-0.0036$ & 6, 1.00 & $-0.0146$ & 5, .557 \\
    \addlinespace[2pt]
    Computers & DEMO & $+0.0873$ & 10, .002 & $+0.1681$ & 10, .002 & $+0.0537$ & 10, .002 & $+0.0979$ & 10, .002 \\
    & NSReg & $+0.0183$ & 9, .014 & $+0.0716$ & 10, .002 & $-0.0173$ & 2, .065 & $+0.0110$ & 6, .432 \\
    \addlinespace[2pt]
    CS & DEMO & $+0.0224$ & 10, .002 & $+0.0540$ & 10, .002 & $+0.0433$ & 10, .002 & $+0.0943$ & 10, .002 \\
    & NSReg & $+0.0161$ & 10, .002 & $+0.0306$ & 10, .002 & $+0.0106$ & 8, .019 & $+0.0159$ & 6, .275 \\
    \addlinespace[2pt]
    ogbn-arxiv & DEMO & $+0.0006$ & 7, .625 & $+0.0021$ & 7, .375 & $+0.0015$ & 6, .375 & $+0.0034$ & 9, .004 \\
    & NSReg & $-0.0317$ & 0, .002 & $-0.0052$ & 2, .037 & $-0.0332$ & 0, .002 & $-0.0076$ & 0, .002 \\
    \bottomrule
  \end{tabular}
\end{table}

\begin{table}[H]
  \centering
  \caption{Where each method is placed on each graph by mean AUC-ROC over all test anomalies, under each rule: 1 is best, 3 is worst. The underlying means are in Table~\ref{tab:main}; only the order is given here, because the order is what the selection rule changes. It changes on Photo and on Computers, and nowhere else.}
  \label{tab:rankall}
  \small
  \begin{tabular}{@{}lcccccccc@{}}
    \toprule
    & \multicolumn{3}{c}{Best epoch on test} & \multicolumn{3}{c}{Best epoch on validation} & \\
    \cmidrule(lr){2-4}\cmidrule(lr){5-7}
    Graph & DEMO & NSReg & OUTPOST & DEMO & NSReg & OUTPOST & Order changes? \\
    \midrule
    Photo & 1 & 2 & 3 & 3 & 1 & 2 & \textbf{yes} \\
    Computers & 3 & 2 & 1 & 3 & 1 & 2 & \textbf{yes} \\
    CS & 3 & 2 & 1 & 3 & 2 & 1 & no \\
    Yelp & 3 & 2 & 1 & 3 & 2 & 1 & no \\
    Amazon & 2 & 1 & 3 & 2 & 1 & 3 & no \\
    T-Finance & 2 & 1 & 3 & 2 & 1 & 3 & no \\
    ogbn-arxiv & 3 & 1 & 2 & 3 & 1 & 2 & no \\
    \bottomrule
  \end{tabular}
\end{table}

\begin{table}[H]
  \centering
  \caption{The same placings computed over unseen classes only, on the four graphs that have unseen classes and on which all three methods run. The rule decides the order on two of the four, the same two as in Table~\ref{tab:rankall}.}
  \label{tab:rankunseen}
  \small
  \begin{tabular}{@{}lcccccccc@{}}
    \toprule
    & \multicolumn{3}{c}{Best epoch on test} & \multicolumn{3}{c}{Best epoch on validation} & \\
    \cmidrule(lr){2-4}\cmidrule(lr){5-7}
    Graph & DEMO & NSReg & OUTPOST & DEMO & NSReg & OUTPOST & Order changes? \\
    \midrule
    Photo & 1 & 2 & 3 & 3 & 1 & 2 & \textbf{yes} \\
    Computers & 3 & 2 & 1 & 3 & 1 & 2 & \textbf{yes} \\
    CS & 3 & 2 & 1 & 3 & 2 & 1 & no \\
    ogbn-arxiv & 3 & 1 & 2 & 3 & 1 & 2 & no \\
    \bottomrule
  \end{tabular}
\end{table}

\section{Oracle Bonus, Budget, and Tuning}\label{app:bonus}

\paragraph{Oracle bonus.} Table~\ref{tab:inflation} gives the oracle bonus of OUTPOST on every graph, over all anomalies and over unseen classes only. Table~\ref{tab:variance} splits the gain of the full model over the variant without pseudo-labeling and the gate into the part that survives the validation rule and the part only the oracle rule collects.

\begin{table}[H]
  \centering
  \caption{The oracle bonus of OUTPOST on each graph, the gain from reading the test metric at the best epoch on test rather than at the best epoch on validation, as mean and standard deviation over ten seeds. Over all anomalies, the bonus is 3 to 35 times larger on the semi-synthetic graphs than on the real ones, and in AUC-ROC it is larger still on unseen classes.}
  \label{tab:inflation}
  \small
  \begin{tabular}{@{}lcccc@{}}
    \toprule
    & \multicolumn{2}{c}{All anomalies} & \multicolumn{2}{c}{Unseen classes only} \\
    \cmidrule(lr){2-3}\cmidrule(lr){4-5}
    Graph & AUC-ROC & AUC-PR & AUC-ROC & AUC-PR \\
    \midrule
    \multicolumn{5}{@{}c}{\emph{Semi-synthetic}} \\
    Photo & $+0.0789$\,{\scriptsize$\pm$.0374} & $+0.0580$\,{\scriptsize$\pm$.0270} & $+0.1779$\,{\scriptsize$\pm$.0769} & $+0.0844$\,{\scriptsize$\pm$.0301} \\
    Computers & $+0.0795$\,{\scriptsize$\pm$.0168} & $+0.1197$\,{\scriptsize$\pm$.0360} & $+0.0972$\,{\scriptsize$\pm$.0195} & $+0.1617$\,{\scriptsize$\pm$.0435} \\
    CS & $+0.0453$\,{\scriptsize$\pm$.0144} & $+0.0895$\,{\scriptsize$\pm$.0250} & $+0.0506$\,{\scriptsize$\pm$.0155} & $+0.1113$\,{\scriptsize$\pm$.0276} \\
    \addlinespace[2pt]
    \multicolumn{5}{@{}c}{\emph{OGB}} \\
    ogbn-arxiv & $+0.0171$\,{\scriptsize$\pm$.0083} & $+0.0101$\,{\scriptsize$\pm$.0055} & $+0.0441$\,{\scriptsize$\pm$.0138} & $+0.0209$\,{\scriptsize$\pm$.0082} \\
    ogbn-mag & $+0.0204$\,{\scriptsize$\pm$.0025} & $+0.0011$\,{\scriptsize$\pm$.0006} & $+0.0258$\,{\scriptsize$\pm$.0030} & $+0.0005$\,{\scriptsize$\pm$.0001} \\
    \addlinespace[2pt]
    \multicolumn{5}{@{}c}{\emph{Real}} \\
    Yelp & $+0.0023$\,{\scriptsize$\pm$.0023} & $+0.0119$\,{\scriptsize$\pm$.0137} & -- & -- \\
    Amazon & $+0.0114$\,{\scriptsize$\pm$.0097} & $+0.0210$\,{\scriptsize$\pm$.0233} & -- & -- \\
    T-Finance & $+0.0141$\,{\scriptsize$\pm$.0077} & $+0.0375$\,{\scriptsize$\pm$.0191} & -- & -- \\
    \bottomrule
  \end{tabular}
\end{table}

\begin{table}[H]
  \centering
  \caption{The advantage of full OUTPOST over the variant with pseudo-labeling and the gate removed, split into the part that survives the validation rule and the part only the oracle rule collects. On the semi-synthetic graphs, 43 to 62 percent of the apparent benefit is of the second kind. Runs are at each graph's development budget (200 epochs on Photo, Computers, and CS; 400 elsewhere).}
  \label{tab:variance}
  \small
  \begin{tabular}{@{}lrcccrc@{}}
    \toprule
    Graph & Seeds & Total advantage & Survives validation & Collected by the oracle & Share & $p$ \\
    \midrule
    Photo & 10 & $+0.0382$ & $+0.0147$ & $+0.0235$ & 62\% & 0.193 \\
    Computers & 5 & $+0.0645$ & $+0.0365$ & $+0.0280$ & 43\% & 0.062 \\
    CS & 5 & $+0.0648$ & $+0.0362$ & $+0.0285$ & 44\% & 0.062 \\
    Yelp & 10 & $-0.0058$ & $-0.0067$ & $+0.0008$ & -- & 0.027 \\
    Amazon & 10 & $+0.0023$ & $+0.0024$ & $-0.0001$ & -- & 0.770 \\
    \bottomrule
  \end{tabular}
\end{table}

\paragraph{Training budget.}\label{app:budget} Table~\ref{tab:budget} compares 200 and 400 epochs. On Photo, the two methods tie at 200 epochs, and DEMO is ahead at 400; 77\% of DEMO's gap to its published Photo value closes with budget alone (Table~\ref{tab:repro}). We chose 400 for all methods because the baseline is properly trained there, although it costs OUTPOST the Photo tie.

\begin{table}[H]
  \centering
  \caption{OUTPOST against DEMO at DEMO's released budget of 200 epochs and at the 400 epochs used throughout, scored at the best epoch on test and paired by seed. On Photo, the two tie at 200 epochs, and DEMO is ahead at 400.}
  \label{tab:budget}
  \small
  \begin{tabular}{@{}lrrccrcrc@{}}
    \toprule
    & & & & & \multicolumn{2}{c}{AUC-ROC} & \multicolumn{2}{c}{AUC-PR} \\
    \cmidrule(lr){6-7}\cmidrule(lr){8-9}
    Graph & Epochs & Seeds & OUTPOST & DEMO & $\Delta$ & Wins, $p$ & $\Delta$ & Wins, $p$ \\
    \midrule
    Photo & 200 & 10 & 0.8374 & 0.8403 & $-0.0029$ & 3, .557 & $-0.0022$ & 3, .557 \\
    Photo & 400 & 10 & 0.8703 & 0.8879 & $-0.0176$ & 3, .065 & $-0.0291$ & 1, .037 \\
    \addlinespace[2pt]
    Computers & 200 & 5 & 0.8214 & 0.7685 & $+0.0529$ & 5, .062 & $+0.0880$ & 5, .062 \\
    Computers & 400 & 10 & 0.8510 & 0.7782 & $+0.0728$ & 10, .002 & $+0.1232$ & 10, .002 \\
    \addlinespace[2pt]
    CS & 200 & 5 & 0.9832 & 0.9603 & $+0.0228$ & 5, .062 & $+0.0469$ & 5, .062 \\
    CS & 400 & 10 & 0.9842 & 0.9646 & $+0.0196$ & 10, .002 & $+0.0419$ & 10, .002 \\
    \bottomrule
  \end{tabular}
\end{table}

\paragraph{Hyperparameter selection.} Table~\ref{tab:hparam} gives the validation-only tuning of OUTPOST and Table~\ref{tab:bsel} the same rule applied to NSReg. Table~\ref{tab:selsurvive} re-tests at ten seeds the three selections that departed from the default, and Table~\ref{tab:selnoise} measures the noise in the statistic the rule reads.

\begin{table}[H]
  \centering
  \caption{Validation-only tuning of OUTPOST, using three selection seeds and a tie band of 0.002. All arms are averaged over the same three selection seeds, so values differ from the ten-seed means in Table~\ref{tab:main}. The last column is what choosing the best arm on the test set instead would have added; contrary to registered prediction P30, it does not grow with the training budget.}
  \label{tab:hparam}
  \small
  \begin{tabular}{@{}lrrcccc@{}}
    \toprule
    & & & Default & \multicolumn{2}{c}{Test AUC-ROC of the arm chosen by} & \\
    \cmidrule(lr){5-6}
    Graph & Epochs & Arms & kept & validation & the test set & Difference \\
    \midrule
    Photo & 200 & 14 & yes & 0.8367 & 0.8532 & $+0.0166$ \\
    Photo & 400 & 14 & yes & 0.8703 & 0.8867 & $+0.0164$ \\
    Computers & 200 & 11 & yes & 0.8193 & 0.8261 & $+0.0068$ \\
    Computers & 400 & 11 & yes & 0.8485 & 0.8527 & $+0.0042$ \\
    CS & 200 & 11 & yes & 0.9838 & 0.9841 & $+0.0003$ \\
    Yelp & 400 & 14 & yes & 0.7580 & 0.7618 & $+0.0038$ \\
    Amazon & 400 & 14 & no & 0.9406 & 0.9511 & $+0.0105$ \\
    \bottomrule
  \end{tabular}
\end{table}

\begin{table}[H]
  \centering
  \caption{The same tuning rule applied to NSReg over a grid of learning rate against weight decay, at three seeds. CS is excluded because the sweep was too expensive there. Where the rule leaves the released configuration, it makes the baseline stronger at three seeds; Table~\ref{tab:selsurvive} re-tests these selections at ten.}
  \label{tab:bsel}
  \small
  \begin{tabular}{@{}lccccccc@{}}
    \toprule
    & Released & Arms inside & Validation & \multicolumn{2}{c}{Validation AUC-ROC} & \multicolumn{2}{c}{Change on test} \\
    \cmidrule(lr){5-6}\cmidrule(lr){7-8}
    Graph & kept & the tie band & spread & released & selected & oracle & validation \\
    \midrule
    Photo & no & 1 of 4 & 0.0037 & 0.8573 & 0.8679 & $+0.0106$ & $+0.0118$ \\
    Computers & yes & 4 of 4 & 0.0008 & 0.8339 & 0.8339 & $+0.0000$ & $+0.0000$ \\
    Yelp & yes & 1 of 4 & 0.0058 & 0.7488 & 0.7488 & $+0.0000$ & $+0.0000$ \\
    Amazon & no & 2 of 4 & 0.0044 & 0.9553 & 0.9600 & $+0.0047$ & $+0.0225$ \\
    \bottomrule
  \end{tabular}
\end{table}

\begin{table}[H]
  \centering
  \caption{The three tuning decisions that departed from the default and were then re-tested at ten seeds. A margin is a difference in validation AUC-ROC between the selected and the default arm; the test column is the same difference measured on test, paired by seed. Wins are seeds, out of ten, on which the selected arm is ahead. Two of the three reverse or vanish once the seed count is raised.}
  \label{tab:selsurvive}
  \small
  \begin{tabular}{@{}llccccl@{}}
    \toprule
    Method & Graph & \multicolumn{2}{c}{Validation margin} & Test & Wins, $p$ & Outcome \\
    \cmidrule(lr){3-4}
    & & 3 seeds & 10 seeds & difference & & \\
    \midrule
    OUTPOST & Amazon & $+0.0057$ & $-0.0004$ & $-0.0040$ & 1, .004 & reverses \\
    NSReg & Photo & $+0.0037$ & $+0.0008$ & $-0.0009$ & 4, .625 & falls inside the tie band \\
    NSReg & Amazon & $+0.0044$ & $+0.0057$ & $+0.0054$ & 9, .006 & holds, and was adopted \\
    \bottomrule
  \end{tabular}
\end{table}

\begin{table}[H]
  \centering
  \caption{Noise in the statistic that the tuning rule reads. For each graph, we take every pair of arms of one method that share seeds, and report the spread of their paired difference in validation AUC-ROC. The last column is the number of seeds at which that spread would fall inside the 0.002 tie band. T-Finance rests on 15 pairs and is an order-of-magnitude estimate only.}
  \label{tab:selnoise}
  \small
  \begin{tabular}{@{}lrcccr@{}}
    \toprule
    & & Standard deviation & \multicolumn{2}{c}{Standard error at} & Seeds needed \\
    \cmidrule(lr){4-5}
    Graph & Arm pairs & of the difference & 3 seeds & 10 seeds & for 0.002 \\
    \midrule
    Photo & 1{,}076 & 0.0073 & 0.0042 & 0.0023 & 14 \\
    Computers & 371 & 0.0077 & 0.0044 & 0.0024 & 15 \\
    CS & 140 & 0.0094 & 0.0054 & 0.0030 & 23 \\
    Yelp & 633 & 0.0134 & 0.0077 & 0.0042 & 45 \\
    Amazon & 217 & 0.0066 & 0.0038 & 0.0021 & 11 \\
    T-Finance & 15 & 0.0141 & 0.0082 & 0.0045 & 51 \\
    \bottomrule
  \end{tabular}
\end{table}

With OUTPOST's Amazon row moved to the arm that validation selected, its oracle AUC-ROC is 0.9494 $\pm$ 0.0076 instead of 0.9534, and its deficit to DEMO grows from $-0.0021$ to $-0.0061$ ($p=0.037$). We had registered both outcomes (P32, P33). The main table keeps the default arm, because the preference that selected the other arm does not exist at ten seeds (P40); neither choice changes which method leads Amazon. Under the same rule, both baselines gain on Amazon ($+0.0054$ for NSReg, $+0.0041$ for DEMO at three seeds) and only OUTPOST loses (P37).

\section{Ablations}\label{app:ablations}

\begin{table}[H]
  \centering
  \caption{Every ablation we ran, paired by seed against the full model at each graph's development budget (200 epochs on Photo, Computers, and CS; 400 elsewhere), as the variant minus the full model. A row that says ``on'' switches on a component that is off by default; HopMix and the spectral gate are the two components we tested and rejected. Wins are seeds on which the variant is ahead, out of the stated number; the only ties are two Yelp conformal cells. At five paired seeds, the smallest attainable $p$ is 0.062.}
  \label{tab:ablfull}
  \small
  \setlength{\tabcolsep}{4pt}
  \begin{tabular}{@{}llrrcrc@{}}
    \toprule
    & & & \multicolumn{2}{c}{AUC-ROC} & \multicolumn{2}{c}{AUC-PR} \\
    \cmidrule(lr){4-5}\cmidrule(lr){6-7}
    Graph & Variant & Seeds & $\Delta$ & Wins, $p$ & $\Delta$ & Wins, $p$ \\
    \midrule
    Photo & atlas gate off & 5 & $+0.0003$ & 2, 1.00 & $-0.0002$ & 2, 1.00 \\
    Photo & pseudo-labeling off & 5 & $-0.0377$ & 0, .062 & $-0.0192$ & 0, .062 \\
    Photo & conformal $\to$ fixed 0.95 & 5 & $-0.0037$ & 2, .312 & $+0.0028$ & 3, .438 \\
    Photo & gate and pseudo-labeling off & 10 & $-0.0381$ & 0, .002 & $-0.0223$ & 1, .004 \\
    Photo & anomaly synthesis off & 5 & $+0.0183$ & 5, .062 & $+0.0074$ & 4, .188 \\
    Photo & SimSample on & 5 & $-0.0586$ & 0, .062 & $-0.0321$ & 0, .062 \\
    \addlinespace[2pt]
    Computers & atlas gate off & 5 & $-0.0012$ & 3, 1.00 & $+0.0010$ & 4, .438 \\
    Computers & pseudo-labeling off & 5 & $-0.0645$ & 0, .062 & $-0.1003$ & 0, .062 \\
    Computers & conformal $\to$ fixed 0.95 & 5 & $-0.0366$ & 0, .062 & $-0.0647$ & 0, .062 \\
    Computers & gate and pseudo-labeling off & 5 & $-0.0645$ & 0, .062 & $-0.1003$ & 0, .062 \\
    Computers & SimSample on & 5 & $+0.0053$ & 3, .312 & $-0.0131$ & 0, .062 \\
    \addlinespace[2pt]
    CS & atlas gate off & 5 & $+0.0000$ & 3, .812 & $-0.0001$ & 2, .312 \\
    CS & pseudo-labeling off & 5 & $-0.0648$ & 0, .062 & $-0.1245$ & 0, .062 \\
    CS & conformal $\to$ fixed 0.95 & 5 & $-0.0047$ & 0, .062 & $-0.0084$ & 0, .062 \\
    CS & gate and pseudo-labeling off & 5 & $-0.0648$ & 0, .062 & $-0.1245$ & 0, .062 \\
    CS & SimSample on & 5 & $+0.0005$ & 4, .125 & $+0.0025$ & 5, .062 \\
    \addlinespace[2pt]
    Yelp & atlas gate off & 10 & $+0.0005$ & 5, .625 & $+0.0011$ & 6, .625 \\
    Yelp & pseudo-labeling off & 10 & $+0.0059$ & 9, .027 & $+0.0038$ & 7, .232 \\
    Yelp & conformal $\to$ fixed 0.95 & 5 & $-0.0002$ & 0, .180 & $-0.0027$ & 0, .068 \\
    Yelp & gate and pseudo-labeling off & 10 & $+0.0059$ & 8, .027 & $+0.0041$ & 7, .193 \\
    Yelp & SimSample off & 10 & $-0.0263$ & 0, .002 & $-0.0504$ & 0, .002 \\
    Yelp & SimSample, shuffled order (placebo) & 10 & $-0.0291$ & 0, .002 & $-0.0555$ & 0, .002 \\
    Yelp & add spectral gate & 5 & $+0.0035$ & 5, .062 & $+0.0035$ & 3, .438 \\
    Yelp & add HopMix fusion & 5 & $-0.0012$ & 3, .812 & $-0.0116$ & 1, .188 \\
    Yelp & add anomaly synthesis & 5 & $-0.0016$ & 2, .812 & $-0.0318$ & 1, .188 \\
    \addlinespace[2pt]
    Amazon & pseudo-labeling off & 10 & $-0.0018$ & 3, .432 & $-0.0057$ & 2, .105 \\
    Amazon & atlas gate off & 10 & $-0.0002$ & 3, .322 & $-0.0001$ & 5, 1.00 \\
    Amazon & conformal $\to$ fixed 0.95 & 10 & $+0.0009$ & 4, 1.00 & $-0.0012$ & 5, .769 \\
    Amazon & gate and pseudo-labeling off & 10 & $-0.0023$ & 4, .375 & $-0.0061$ & 2, .084 \\
    Amazon & SimSample, shuffled order (placebo) & 10 & $-0.0091$ & 1, .006 & $-0.0157$ & 0, .002 \\
    Amazon & SimSample on & 10 & $-0.0059$ & 1, .014 & $-0.0281$ & 0, .002 \\
    \addlinespace[2pt]
    T-Finance & pseudo-labeling off & 10 & $+0.0097$ & 9, .004 & $+0.0349$ & 9, .004 \\
    T-Finance & atlas gate off & 10 & $-0.0000$ & 4, .922 & $-0.0024$ & 4, .695 \\
    T-Finance & conformal $\to$ fixed 0.95 & 10 & $+0.0027$ & 8, .131 & $+0.0086$ & 8, .105 \\
    T-Finance & SimSample on & 10 & $+0.0128$ & 9, .004 & $+0.0774$ & 10, .002 \\
    \addlinespace[2pt]
    ogbn-arxiv & SimSample on & 5 & $-0.0004$ & 2, .625 & $+0.0008$ & 4, .188 \\
    \bottomrule
  \end{tabular}
\end{table}

\begin{table}[H]
  \centering
  \caption{The atlas gate removed, at the 400-epoch protocol of the main results; $\Delta$ is off minus on. At the best epoch on test, nothing approaches significance, and the largest movement is 0.0016. Under the validation rule, removing the gate costs 0.0014 on CS ($p=0.037$); no other cell is significant.}
  \label{tab:gateoff}
  \small
  \begin{tabular}{@{}lrcccccc@{}}
    \toprule
    & & \multicolumn{3}{c}{Best epoch on test} & \multicolumn{3}{c}{Best epoch on validation} \\
    \cmidrule(lr){3-5}\cmidrule(lr){6-8}
    Graph & Seeds & on & off & $\Delta$, wins, $p$ & on & off & $\Delta$, wins, $p$ \\
    \midrule
    Photo & 10 & 0.8703 & 0.8701 & $-0.0002$, 6, .769 & 0.7914 & 0.7860 & $-0.0053$, 4, .492 \\
    Computers & 10 & 0.8510 & 0.8495 & $-0.0016$, 5, .625 & 0.7715 & 0.7709 & $-0.0006$, 3, .492 \\
    CS & 10 & 0.9842 & 0.9842 & $+0.0001$, 6, .375 & 0.9389 & 0.9374 & $-0.0014$, 1, .037 \\
    Yelp & 10 & 0.7448 & 0.7454 & $+0.0005$, 5, .625 & 0.7425 & 0.7418 & $-0.0007$, 3, .322 \\
    Amazon & 10 & 0.9534 & 0.9532 & $-0.0002$, 3, .322 & 0.9420 & 0.9464 & $+0.0043$, 7, .105 \\
    T-Finance & 10 & 0.9087 & 0.9087 & $-0.0000$, 4, .922 & 0.8946 & 0.8980 & $+0.0034$, 7, .105 \\
    ogbn-arxiv & 5 & 0.6237 & 0.6225 & $-0.0012$, 1, .188 & 0.6048 & 0.6048 & $-0.0001$, 3, 1.00 \\
    \bottomrule
  \end{tabular}
\end{table}

Removing the gate and pseudo-labeling together equals removing pseudo-labeling alone to four decimals on Computers and CS, so the two do not interact; we had registered that the joint removal would cost more than the sum of the two (P20, falsified). Anomaly synthesis has mixed effects: removing it raises Photo AUC-ROC by 0.018 on five of five seeds, and adding it on Yelp lowers AUC-PR by 0.032. We keep the configuration fixed per graph kind (Section~\ref{sec:method}) rather than switching components per graph, which would add tuning decisions the protocol must account for.

\section{SimSample}\label{app:simsample}

\begin{figure}[H]
  \centering
  \includegraphics[width=0.81\linewidth]{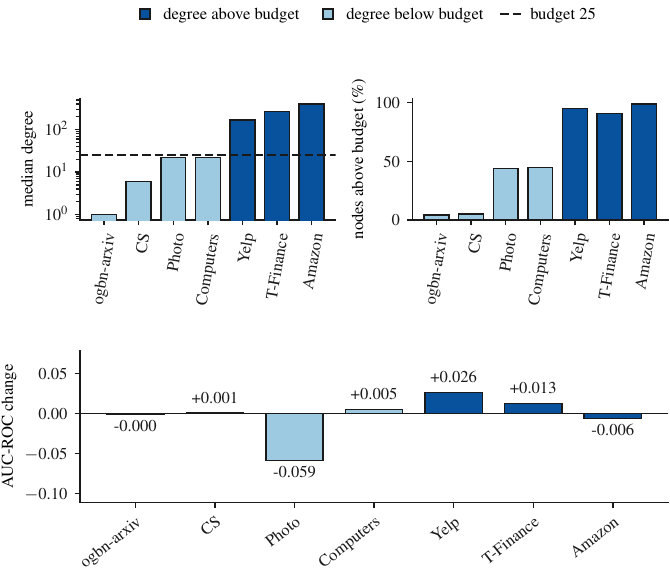}
  \caption{The registered degree-budget account of SimSample, tested against the data. Graphs are ordered by median degree. The upper left panel gives that degree against the first-hop sampling budget of 25, the upper right the share of nodes whose degree exceeds the budget, and the wide panel below the AUC-ROC that SimSample adds or removes, with the paired seed count behind each value given in Table~\ref{tab:simsample}. Color means the same thing in all three panels: whether the graph sits above or below the budget. Were the budget the mechanism, every dark bar in the lower panel would be positive and every light one negative. Two of the three graphs above the budget improve, and Amazon does not, and of the four below it, only Photo is clearly harmed.}
  \label{fig:regime}
\end{figure}

This appendix reports the evidence behind Section~\ref{sec:rq5}: the by-graph comparison in Table~\ref{tab:simsample}, the registered regime account in Fig.~\ref{fig:regime}, and the budget intervention below.

\begin{table}[H]
  \centering
  \caption{SimSample switched on minus switched off, by graph, ordered by median degree; wins are out of the stated seeds. We had registered that it would help where node degree exceeds the first-hop budget of 25 and hurt below it. Photo, Yelp, and T-Finance fit that account; CS and ogbn-arxiv, which are below the budget but unaffected, and Amazon, the densest graph, which is harmed, do not.}
  \label{tab:simsample}
  \small
  \begin{tabular}{@{}lrrrcccc@{}}
    \toprule
    & Median & Nodes above & & \multicolumn{2}{c}{AUC-ROC} & \multicolumn{2}{c}{AUC-PR} \\
    \cmidrule(lr){5-6}\cmidrule(lr){7-8}
    Graph & degree & the budget & Seeds & $\Delta$ & Wins & $\Delta$ & Wins \\
    \midrule
    ogbn-arxiv & 1 & 4.3\% & 5 & $-0.0004$ & 2 & $+0.0008$ & 3 \\
    CS & 6 & 5.1\% & 5 & $+0.0005$ & 4 & $+0.0025$ & 5 \\
    Photo & 22 & 43.8\% & 5 & $-0.0586$ & 0 & $-0.0321$ & 0 \\
    Computers & 22 & 45.0\% & 5 & $+0.0053$ & 3 & $-0.0131$ & 0 \\
    Yelp & 168 & 95.2\% & 10 & $+0.0263$ & 10 & $+0.0504$ & 10 \\
    T-Finance & 265 & 90.8\% & 10 & $+0.0128$ & 9 & $+0.0774$ & 10 \\
    Amazon & 403 & 98.7\% & 10 & $-0.0059$ & 1 & $-0.0281$ & 0 \\
    \bottomrule
  \end{tabular}
\end{table}

\textbf{Budget intervention on Yelp.} With first-hop budgets 25, 50, 100, and 200, AUC-PR with SimSample is 0.3824, 0.3816, 0.3676, and 0.3624, and without it 0.3320, 0.3422, 0.3407, and 0.3407. The ``off'' arm rises by 0.010 from budget 25 to 50 and is flat thereafter, while the ``on'' arm falls by 0.020 across the sweep; most of the shrinkage therefore comes from the ``on'' arm, and the sweep mainly measures neighbor selection rather than receptive-field size. We had predicted that the gain would vanish at budget 200, above the median degree of 168 (P11). It did not ($+0.0217$, five of five seeds), because 37\% of Yelp's nodes still exceed that budget. A two-term model fitted afterward predicted $-0.083$ on ogbn-arxiv; the measured value is $+0.0008$, and we discard the model (P13).

\section{Training-Free Diagnostic}\label{app:diag}

\begin{figure}[H]
  \centering
  \includegraphics[width=0.44\linewidth]{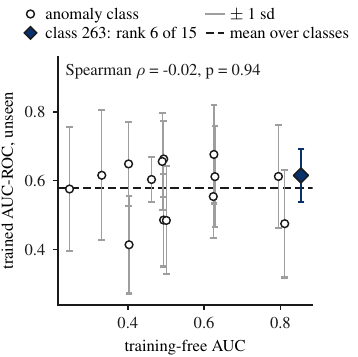}
  \caption{The 15 ogbn-mag classes, each placed by the AUC, the training-free diagnostic gives it without propagation, against the AUC trained OUTPOST reaches on that same class when the class is held out as unseen. Bars are the standard deviation over rotations and seeds, and the dashed line is the mean over classes. The two quantities do not rank together. The class the diagnostic rates easiest ranks sixth of fifteen once a model is trained, and what distinguishes it instead is an unusually small spread, so the diagnostic says more about how stable a class is than about how detectable it is.}
  \label{fig:p8}
\end{figure}

The diagnostic uses no trained model. For propagation depth $k\in\{0,1,2,3\}$ we compute $\hat A^k X$ with the symmetric normalized adjacency $\hat A$, fit spherical $k$-means prototypes on 5\% of the normal nodes, score every node by its distance to the nearest prototype, and compute one AUC-ROC per anomaly class (Table~\ref{tab:diagclasses}). ``Best prototype AUC'' is the maximum over $k$, and ``propagation gain'' is the AUC at $k=3$ minus the AUC at $k=0$. The correlation reported in Section~\ref{sec:rq6} describes this diagnostic; it does not show that the diagnostic predicts a trained model, and Figure~\ref{fig:p8} shows that on ogbn-mag it does not.

\begin{figure}[H]
  \centering
    \begin{minipage}[b]{0.655\linewidth}\centering
    \includegraphics[width=\linewidth]{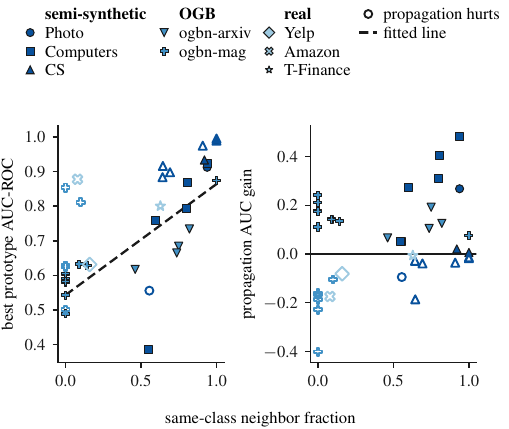}
  \end{minipage}\hfill
  \begin{minipage}[b]{0.325\linewidth}\centering
    \raisebox{8.6pt}{\includegraphics[width=\linewidth]{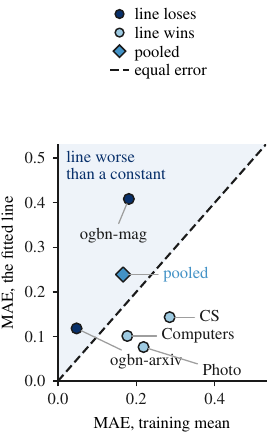}}
  \end{minipage}
  \caption{\textbf{Left and middle:} all 37 anomaly classes, placed by how clustered the class is, meaning the median fraction of a class node's neighbors that carry the same label. The left panel gives the AUC of the training-free diagnostic reaches on that class, with the dashed line fitted to the classes that propagation helps; the middle panel shows how much AUC the class gains once features are propagated. Each marker shape is one graph, the color says how the anomalies were produced, and a hollow marker means propagation hurt that class. \textbf{Right:} what the fitted line is worth for classes it was not fitted on. Each point holds out one graph and compares the error of the line against the error of simply predicting the training mean, so a point above the diagonal is a graph that the line predicts worse than a constant does. Both OGB graphs and the pooled error fall there. The relation describes the classes it was fitted to and does not predict new ones.}
  \label{fig:law}
\end{figure}

\textbf{Relation to homophily.} Across 37 classes in eight graphs, the best prototype AUC correlates with the class's median same-class neighbor fraction at Spearman $\rho=0.778$, with a within-graph permutation $p=5\times10^{-5}$ and a bootstrap interval over graphs of $[0.39,0.91]$. Among the 21 classes that propagation helps, the correlation is $0.896$; clustered classes gain up to $0.48$ AUC from propagation, and scattered classes lose (Fig.~\ref{fig:law}, left and middle).

\textbf{Out-of-sample tests.} The relation does not hold up as a predictor. Registered before any training on Amazon, the fitted line predicted a ceiling of $0.615$, whereas the diagnostic alone reached $0.878$ and the trained model $0.953$ (P23, falsified). Held out one graph at a time, the line predicts unseen classes worse than a constant does, with mean absolute error $0.239$ against $0.166$ (Table~\ref{tab:lawcv}; Fig.~\ref{fig:law}, right). On the 15 ogbn-mag classes, the diagnostic has no rank correlation with the AUC a trained model attains on the same class ($\rho=-0.02$, $p=0.94$), and the class it rates easiest ranks sixth (P8, falsified; Fig.~\ref{fig:p8}). A closed-form crossover derived from a signal-to-noise argument agreed with the data on 8 of 16 classes, which is chance (Appendix~\ref{app:theory}). We therefore report the relation as a description of these graphs, not as a predictor of new ones.

\begin{table}[H]
  \centering
  \caption{The training-free diagnostic on all 37 anomaly classes. Same-class fraction is the median share of a class node's neighbors carrying the same label. Best AUC is the highest AUC-ROC, the diagnostic reaches over propagation depths zero to three, and gain is the AUC at depth three minus the AUC at depth zero, so a negative value means propagation made the class harder to find. The 37 classes are split into two columns to fit the page.}
  \label{tab:diagclasses}
  \small
  \setlength{\tabcolsep}{4pt}
  \begin{tabular}{@{}lrccc@{\hspace{14pt}}lrccc@{}}
    \toprule
    Graph & Class & Frac. & Best AUC & Gain & Graph & Class & Frac. & Best AUC & Gain \\
    \midrule
    Photo & 0 & 0.938 & 0.912 & $+0.268$ & ogbn-arxiv & 8 & 0.750 & 0.684 & $+0.192$ \\
    Photo & 7 & 0.556 & 0.556 & $-0.095$ & ogbn-arxiv & 10 & 0.462 & 0.617 & $+0.066$ \\
    Computers & 0 & 0.800 & 0.793 & $+0.310$ & ogbn-arxiv & 34 & 0.737 & 0.665 & $+0.105$ \\
    Computers & 3 & 0.597 & 0.759 & $+0.271$ & ogbn-mag & 2 & 0.000 & 0.580 & $+0.177$ \\
    Computers & 5 & 0.940 & 0.924 & $+0.482$ & ogbn-mag & 39 & 0.143 & 0.629 & $+0.135$ \\
    Computers & 6 & 0.550 & 0.384 & $+0.053$ & ogbn-mag & 143 & 1.000 & 0.874 & $+0.076$ \\
    Computers & 9 & 0.806 & 0.867 & $+0.405$ & ogbn-mag & 151 & 0.000 & 0.591 & $+0.111$ \\
    CS & 0 & 1.000 & 0.989 & $-0.017$ & ogbn-mag & 176 & 0.100 & 0.811 & $-0.105$ \\
    CS & 1 & 0.645 & 0.916 & $-0.186$ & ogbn-mag & 195 & 0.000 & 0.582 & $+0.175$ \\
    CS & 3 & 0.909 & 0.975 & $-0.036$ & ogbn-mag & 206 & 0.091 & 0.632 & $+0.142$ \\
    CS & 6 & 0.692 & 0.898 & $-0.039$ & ogbn-mag & 215 & 0.000 & 0.501 & $-0.162$ \\
    CS & 8 & 0.643 & 0.884 & $-0.028$ & ogbn-mag & 216 & 0.000 & 0.626 & $-0.229$ \\
    CS & 9 & 1.000 & 0.996 & $-0.013$ & ogbn-mag & 231 & 0.000 & 0.543 & $+0.211$ \\
    CS & 12 & 1.000 & 0.991 & $+0.006$ & ogbn-mag & 263 & 0.000 & 0.854 & $-0.187$ \\
    CS & 14 & 0.920 & 0.934 & $+0.021$ & ogbn-mag & 282 & 0.000 & 0.605 & $+0.111$ \\
    Yelp & 1 & 0.160 & 0.631 & $-0.081$ & ogbn-mag & 321 & 0.000 & 0.489 & $+0.242$ \\
    Amazon & 1 & 0.080 & 0.878 & $-0.174$ & ogbn-mag & 327 & 0.000 & 0.624 & $-0.168$ \\
    T-Finance & 1 & 0.629 & 0.800 & $-0.006$ & ogbn-mag & 341 & 0.000 & 0.628 & $-0.402$ \\
    ogbn-arxiv & 4 & 0.821 & 0.734 & $+0.126$ & & & & & \\
    \bottomrule
  \end{tabular}
\end{table}

\begin{table}[H]
  \centering
  \caption{Leaving one graph out and predicting its classes from the line fitted to the rest, against the alternative of predicting the training mean. The line wins on the three semi-synthetic graphs and loses on both OGB graphs and on the pooled error, so it describes the classes it was fitted to rather than predicting new ones. The 16 classes that propagation does not help are excluded: for 13 of them, the best AUC equals the AUC without propagation by construction, so that the branch carries no predictive content.}
  \label{tab:lawcv}
  \small
  \begin{tabular}{@{}lrccl@{}}
    \toprule
    & & \multicolumn{2}{c}{Mean absolute error} & \\
    \cmidrule(lr){3-4}
    Held-out graph & Classes & fitted line & training mean & Lower error \\
    \midrule
    Photo & 1 & 0.0763 & 0.2184 & line \\
    Computers & 5 & 0.1011 & 0.1770 & line \\
    CS & 2 & 0.1433 & 0.2850 & line \\
    ogbn-arxiv & 4 & 0.1181 & 0.0475 & training mean \\
    ogbn-mag & 9 & 0.4084 & 0.1809 & training mean \\
    \midrule
    Pooled & 21 & 0.2389 & 0.1663 & training mean \\
    \bottomrule
  \end{tabular}
\end{table}

\section{A Signal-to-Noise Account of Propagation}\label{app:theory}

We tested a simple signal-to-noise account of when propagation helps. Assume each node feature is a class mean plus independent noise of variance $\sigma^2$, and let $\delta$ be the difference between the anomaly mean and the normal mean $m_n$. For an anomaly $v$ with closed-neighborhood size $d_v$ and same-class neighbor fraction $h_v$, one step of mean aggregation gives $\mathbb{E}[x^{(1)}_v]-m_n=h_v\delta$ and $\mathrm{Var}[x^{(1)}_v]=\sigma^2/d_v$. The signal-to-noise ratio therefore changes by the factor $h_v\sqrt{d_v}$, and propagation should help exactly when $h_v>d_v^{-1/2}$. We tested the sign of $h-d^{-1/2}$ against the sign of the measured propagation gain on the 16 classes available at that time. It agreed on 8 of 16, which is a chance. Real degrees are large, so the rule predicts a gain almost everywhere, whereas classes already separable from the features are lost under propagation, regardless of their $h$. Two assumptions fail: neighbor noise is correlated and does not fall as $1/\sqrt{d}$, and AUC has a ceiling. The continuous margin still ranks with the gain (Spearman $0.756$), so the ingredients are reasonable, but the closed form is not supported, and we make no theoretical claim in this paper.

\section{Model Size and Memory}\label{app:efficiency}

\begin{table}[H]
  \centering
  \caption{Trainable parameters counted on the trained models, and peak GPU memory of OUTPOST against DEMO measured in the same process setting. OUTPOST is the smallest of the three on every graph. Against DEMO, the margin is large throughout; against NSReg, it is large only on the three graphs with observed fraud labels, where the input dimension is small.}
  \label{tab:params}
  \small
  \begin{tabular}{@{}lrrrccc@{}}
    \toprule
    & \multicolumn{3}{c}{Trainable parameters} & \multicolumn{2}{c}{OUTPOST relative to} & Peak memory \\
    \cmidrule(lr){2-4}\cmidrule(lr){5-6}
    Graph & OUTPOST & DEMO & NSReg & DEMO & NSReg & vs.\ DEMO \\
    \midrule
    \multicolumn{7}{@{}c}{\emph{Semi-synthetic}} \\
    Photo & 114{,}113 & 725{,}819 & 118{,}274 & 0.157$\times$ & 0.965$\times$ & 0.872$\times$ \\
    Computers & 116{,}929 & 763{,}329 & 121{,}090 & 0.153$\times$ & 0.966$\times$ & 0.811$\times$ \\
    CS & 889{,}793 & 47{,}648{,}399 & 893{,}954 & 0.019$\times$ & 0.995$\times$ & 0.655$\times$ \\
    \addlinespace[2pt]
    \multicolumn{7}{@{}c}{\emph{OGB}} \\
    ogbn-arxiv & 35{,}137 & 68{,}097 & 39{,}298 & 0.516$\times$ & 0.894$\times$ & -- \\
    ogbn-mag & 35{,}137 & -- & -- & -- & -- & -- \\
    \addlinespace[2pt]
    \multicolumn{7}{@{}c}{\emph{Real}} \\
    Yelp & 7{,}361 & 34{,}209 & 27{,}010 & 0.215$\times$ & 0.273$\times$ & 0.441$\times$ \\
    Amazon & 6{,}913 & 32{,}459 & 26{,}114 & 0.213$\times$ & 0.265$\times$ & 0.458$\times$ \\
    T-Finance & 5{,}953 & 29{,}039 & 24{,}194 & 0.205$\times$ & 0.246$\times$ & 0.338$\times$ \\
    \bottomrule
  \end{tabular}
\end{table}

\begin{figure}[H]
  \centering
  \includegraphics[width=0.45\linewidth]{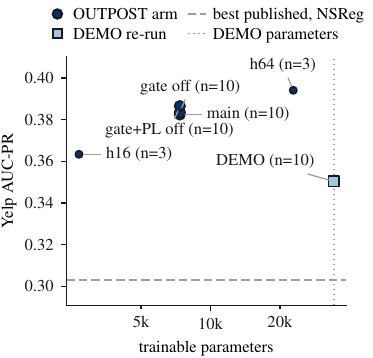}
  \caption{Yelp AUC-PR against trainable parameter count, on a logarithmic axis. Every OUTPOST arm clears both the re-run DEMO and the best published result with at most about a fifth of DEMO's parameters. The arms marked main, gate off, and gate+PL off share a parameter count because the atlas gate is a quantile rule and carries no weights of its own. The two arms at other hidden widths were run for three seeds rather than ten, and are drawn smaller.}
  \label{fig:efficiency}
\end{figure}

Table~\ref{tab:params} gives the parameter counts and peak memory, and Figure~\ref{fig:efficiency} plots the Yelp result against them. The atlas gate is a quantile rule and adds no trainable parameters.

\section{Reporting Checklist}\label{app:checklist}

Seven reporting practices follow from this study, and each is cheap to adopt. (1)~Report the validation-rule score beside the oracle-rule score, for baselines as well as the proposed method. (2)~Re-run the strongest baselines under the same seeds, splits, and budget, and keep transcribed numbers in a separate block. (3)~Choose the seed count so the paired test can reach significance; for the Wilcoxon signed-rank test, the smallest attainable $p$-value is $2^{-(n-1)}$. (4)~Set any tuning tie band against the seed-level noise of the statistic it reads, and re-test a non-default choice at more seeds before adopting it. (5)~Before claiming that a component helps, ablate it on at least one benchmark with observed anomalies. (6)~Distinguish a baseline cell that is infeasible from one that is merely expensive. (7)~Record predictions and their falsifying outcomes before the runs that test them.

\section{Pre-Registration Ledger}\label{app:prereg}

Each prediction was committed to version control before the runs that test it, and Table~\ref{tab:ledger} lists all 40 with their verdicts. P1--P3 belong to the first diagnostic study and were superseded by the eight-graph analysis, and P4 was corrected. Predictions marked (blind) were registered before any run they concern existed; the others were registered before the seeds that test them had been read.

Two predictions concern absolute performance. On ogbn-mag, OUTPOST stays below 0.60 AUC-ROC as registered, but it exceeds the best published value by 0.096 on ten of ten seeds, which we had predicted it would not do; we score P5 as split. On ogbn-arxiv, the result lies inside the published band, as registered (P6).

{\footnotesize
  \begin{longtable}{@{}p{0.04\linewidth}>{\raggedright\arraybackslash}p{0.62\linewidth}>{\raggedright\arraybackslash}p{0.25\linewidth}@{}}
    \caption{All 40 predictions, each committed to version control before the runs that test it, with the outcome. Seventeen were confirmed, twelve falsified, three split, and one null; the remaining seven were superseded, corrected, or hold only in part. Several of the falsified ones concern our own method or our own explanations, and each changed the text of this paper.}\label{tab:ledger}\\
    \toprule
    ID & Prediction & Outcome \\
    \midrule
    \endfirsthead
    \toprule
    ID & Prediction & Outcome \\
    \midrule
    \endhead
    P1 & Anomaly classes differ in same-class fraction, feature dissimilarity, and Dirichlet energy & superseded \\
    P2 & Within a class, same-class fraction correlates with the detection score at the best depth & superseded \\
    P3 & Propagation gain is positive for clustered classes and negative for scattered ones & superseded \\
    P4 & Real fraud (Yelp) has a lower same-class fraction than every relabeled class & corrected: false with ogbn-mag \\
    P5 & ogbn-mag: OUTPOST stays below 0.60 AUC-ROC and does not clearly exceed the published field & split \\
    P6 & ogbn-arxiv: competitive with the published field, not above it by more than seed noise & confirmed \\
    P7 & AUC-ROC over six graphs ranks in the order of the same-class fraction & partially falsified \\
    P8 & ogbn-mag class 263 remains the best-detected class under a trained model & falsified \\
    P9 & Validation-only tuning on Photo cannot separate the arms & confirmed \\
    P10 & Yelp: the SimSample gain decreases monotonically as the sampling budget grows & confirmed \\
    P11 & Yelp: the gain is indistinguishable from zero at budget 200 & falsified \\
    P12 & Yelp: the ``off'' arm does not fall as the budget grows & confirmed \\
    P13 & SimSample is harmful on CS and ogbn-arxiv, more than on Photo & falsified \\
    P14 & CS shows the largest SimSample harm of any graph & falsified \\
    P15 & Validation-only tuning keeps the default on Photo, Computers, CS, and Yelp & confirmed \\
    P16 & Best-on-test minus validation-selected arm is non-trivial on at least one graph & confirmed \\
    P17 & Photo: the SimSample effect is non-monotone in the budget & falsified \\
    P18 & Photo: the effect at budget 100 is smaller in size than at 25 & confirmed in sign only \\
    P19 & Photo: the ``off'' arm does not move with the budget & confirmed \\
    P20 & Removing gate and pseudo-labeling together costs more than the sum of each & falsified \\
    P21 & Removing pseudo-labeling costs more than removing the gate on Computers and CS & confirmed \\
    P22 & Removing the conformal threshold alone stays near zero on Computers and CS & split \\
    P23 & Amazon: the one-line relation predicts a best AUC of 0.615 (band 0.38--0.85) & falsified \\
    P24 & Amazon: feature-visible, achievable AUC near the diagnostic value 0.878 & confirmed in direction \\
    P25 & Amazon: removing pseudo-labeling is clearly positive or clearly negative & null \\
    P26 & T-Finance: diagnostic ceiling within one residual standard deviation of the fitted line & confirmed \\
    P27 & T-Finance: trained OUTPOST lands in $[0.84,0.93]$ AUC-ROC & confirmed \\
    P28 & T-Finance: removing pseudo-labeling is within $\pm0.010$ or positive & confirmed \\
    P29 & 400-epoch sweep: validation keeps the default on Photo and Computers & confirmed \\
    P30 & 400-epoch sweep: the test-selection gap is at least its 200-epoch value & falsified \\
    P31 & Amazon sweep: default kept and test-selection gap below 0.005 & falsified \\
    P32 & Amazon: the validation-selected arm is worse on test than the default at ten seeds & confirmed \\
    P33 & Amazon: the deficit of OUTPOST to DEMO widens under the selected arm & confirmed \\
    P34 & Amazon: removing pseudo-labeling stays within $\pm0.010$ at hidden width 16 & confirmed \\
    P35 & NSReg's validation-selected arms keep their advantage at ten seeds & split: Amazon yes, Photo no \\
    P36 & (blind) DEMO's tuning grid keeps its published configuration on all four graphs & falsified \\
    P37 & (blind) One validation rule moves the three methods in different directions on Amazon & confirmed \\
    P38 & T-Finance: the atlas gate is inert & confirmed \\
    P39 & T-Finance: removing the conformal threshold is negative & falsified \\
    P40 & Amazon: the validation preference for the selected arm survives ten seeds & falsified \\
    \bottomrule
\end{longtable}}

\end{document}

%% file: math_commands.tex
\usepackage{amsmath,amsfonts,bm}

\def\eqref#1{equation~\ref{#1}}

\def\1{\bm{1}}

\DeclareMathAlphabet{\mathsfit}{\encodingdefault}{\sfdefault}{m}{sl}
\SetMathAlphabet{\mathsfit}{bold}{\encodingdefault}{\sfdefault}{bx}{n}

